%% file: main.tex
\documentclass{cta-author}

\usepackage{booktabs} 
\usepackage{graphicx}
\usepackage{epstopdf}
\usepackage{subfigure}
\usepackage{amsmath,amssymb,amsfonts}
\usepackage{xcolor}
\usepackage{enumerate}
\usepackage{amsthm}
\usepackage{diagbox}
\usepackage{enumitem} 
\usepackage{color,colordvi}
\usepackage{textcomp}
\usepackage{epsfig}
\usepackage{url}
\usepackage{mdwlist}
\usepackage{makecell}
\usepackage{verbatim}
\usepackage{multirow,multicol}
\usepackage{setspace}
\usepackage{float}
\usepackage{fancyhdr}
\usepackage[normalem]{ulem}
\usepackage{makecell}
\usepackage{fancyhdr}
\usepackage{flushend}
\usepackage{algorithm}
\usepackage{algorithmic}
\usepackage{natbib}
\usepackage{multirow}
\usepackage{soul}
\usepackage[colorlinks,linkcolor=green,citecolor=green]{hyperref}
{}
{}
{}

\begin{document}
\title{Self-supervised Image Clustering from Multiple Incomplete Views via Constrastive Complementary Generation}

\author{\au{Jiatai Wang$^{1}$}, \au{Zhiwei Xu$^{1,2\corr}$}, \au{Xuewen Yang$^{3}$},  \au{Dongjin Guo$^{1}$},\au{Limin Liu$^{1}$}}

\address{\add{1}{College of Data Science and Application, Inner Mongolia University of Technology, Huhhot, China, 010080}
\add{2}{Institute of Computing Technology, Chinese Academy of Sciences, Beijing, China, 100190}
\add{3}{InnoPeak Technology, Inc, CA, USA, 94303}
\email{xuzhiwei2001@ict.ac.cn}}

\begin{abstract}
Incomplete Multi-View Clustering aims to enhance clustering performance by using data from multiple modalities. Despite the fact that several approaches for studying this issue have been proposed, the following drawbacks still persist: 1) It's difficult to learn latent representations that account for complementarity yet consistency without using label information; 2) and thus fails to take full advantage of the hidden information in incomplete data results in suboptimal clustering performance when complete data is scarce. In this paper, we propose Contrastive Incomplete Multi-View Image Clustering with Generative Adversarial Networks (CIMIC-GAN), which uses GAN to fill in incomplete data and uses double contrastive learning to learn consistency on complete and incomplete data. More specifically, considering diversity and complementary information among multiple modalities, we incorporate autoencoding representation of complete and incomplete data into double contrastive learning to achieve learning consistency. Integrating GANs into the autoencoding process can not only take full advantage of new features of incomplete data, but also better generalize the model in the presence of high data missing rates. Experiments conducted on \textcolor{black}{four} extensively-used datasets show that CIMIC-GAN outperforms state-of-the-art incomplete multi-View clustering methods.
\end{abstract}

\maketitle

\input{Introduction}

\input{Related}
\input{Method}
\input{Experiment}
\input{Conclusion}
\section*{Acknowledgments}
This work  was supported by the National Science Foundation of China (61962045, 62062055, 61650205, 61902382, 61972381), the Open Foundation of Inner Mongolia Key Laboratory of Discipline Inspection and Supervision (IMDBD2020017, IMDBD2020018), the Science and Technology Planning Project of Inner Mongolia Autonomous Region (2019GG372).
\medskip

\bibliography{References}

\end{document}

%% file: Introduction.tex
\section{INTRODUCTION}\label{sec1}

The majority of data in real life is in the form of numerous modalities or multiple views \cite{1}, e.g., the RGB images or depth maps taken by using different types of cameras or from different angles by the same camera. The information in multi-modal data cannot be effectively utilized by the single-modal or single-view method. If we can comprehensively observe different views of the object or use multiple modalities of the image object, we can better build a vision model of the object. Therefore, an effective multi-modal learning approach, particularly one that is unsupervised, is highly desired for real-world vision applications. 
Existing methods all explicitly require multi-view data to satisfy the assumption of cross-view consistency, also known as data completeness, and require that all views of each sample point exist. However,  
considering the scenarios in practice are always missing a lot during data collection or transmission, the \textcolor{black}{data} from complete views are very \textcolor{black}{scarce} and a crucial problem, incomplete multi-view problem (IMP), arises. The key to IMP is whether the missing information can be inferred from existing data, or whether the correct judgment can be made using the available data information.

Self-supervised algorithms of incomplete multi-view clustering (IMC) have been developed to address IMPs in clustering. %by matrix factorization based methods \cite{3,4,5} and GAN based methods \cite{6,7,8}. 
Although significant progress has been made in many fields utilizing existing IMC methods, their performance is limited due to the following drawbacks. Obtaining high-level semantic features with traditional IMC methods is difficult, and degrades clustering performance on complicated real-world data. 
Global latent representations with high-level semantic information should not only promote consistency among views while weakening inconsistency but also account for complementary information. \textcolor{black}{However, some self-supervised IMC methods \cite{3, 4, 5, 8, 46, 48, 49, 50, 52} either extract only the common semantics of different views to maximize inter-view consistency, or only mine the information that complements each other between views by  enhancing subspace learning and fusion, lacking a comprehensive consideration.} Even if some methods \cite{45, 47, 51} take into account consistency and complementary, the clustering performance is still poor due to failure to take full advantage of the hidden information in incomplete data. Many methods \cite{3,4,5,8,21,45,51,52} can only utilize aligned complete data for representation learning based on the consistency assumption of the data. However, %considering the dataset in practice is always missing a lot during collection or transmission,
the \textcolor{black}{data} from complete views are very \textcolor{black}{scarce} and will be not enough to learn consistent information. The information hidden in the incomplete data appears more important for self-supervised IMC. 

Actually, due to the variety and diversity of incomplete data from multiple views, it is quite challenging to utilize incomplete data to its full potential which makes the model more robust in severe scenarios with high missing rates.
To make up for the lack of complete data pairs and comprehensively utilize the hidden information in incomplete data, contrastive learning has been applied in IMC \cite{21, 24, 26, 30, 36}. 
However, while learning consistency from different views, contrastive learning in the feature space will result in the loss of complementary information. We map datasets into multiple-view joint subspaces constructed using auto-encoding representation learning, and then leverage the Generative Adversarial Network (GAN) \cite{40} to fill in the incomplete data to the auto-encoding representation of contrastive learning. 
In this way, the data diversity of each view is completely maintained in our framework by reconstructing its high-dimensional latent representation back to the view, which enables us to contain more complimentary information and use its latent representation to construct a contrastive learning module. 
The above design allows the representations learned by our framework to be complementary yet consistent. 
Ultimately, a new self-supervised IMC framework, named Contrastive Incomplete Multi-View Image Clustering with Generative Adversarial Network (CIMIC-GAN), is proposed. It integrates several view-specific autoencoders, mutually symmetric prediction models, and GAN into a double contrastive complementary learning for self-supervised IMC.

As indicated in Fig. \ref{fig1}, CIMIC-GAN consists of three modules: view reconstruction module with GAN, contrastive consistency module, and contrastive prediction module. 
The view reconstruction module with GAN projects the view data into a specific view latent space, which is also the premise of consistent learning. At the same time, GAN can also complete the missing views to form a representation. The contrastive consistency module and the contrastive prediction module constitute a double contrastive learning framework for consistency learning, which maximizes the mutual information while making the latent representations predict each other. It is worth mentioning that, although the essence of GAN and the contrastive prediction module is distinct, they are both complementary to missing views. The contrastive prediction module can learn mutual representations between complete views. On the other hand, GAN can generate new latent distributions from incomplete view data, allowing the contrastive prediction module to be more 
comprehensively trained. The primary contributions of our work are illustrated as follows:
\begin{figure*}[t]
  \begin{center}
  \includegraphics[width=0.85\textwidth]{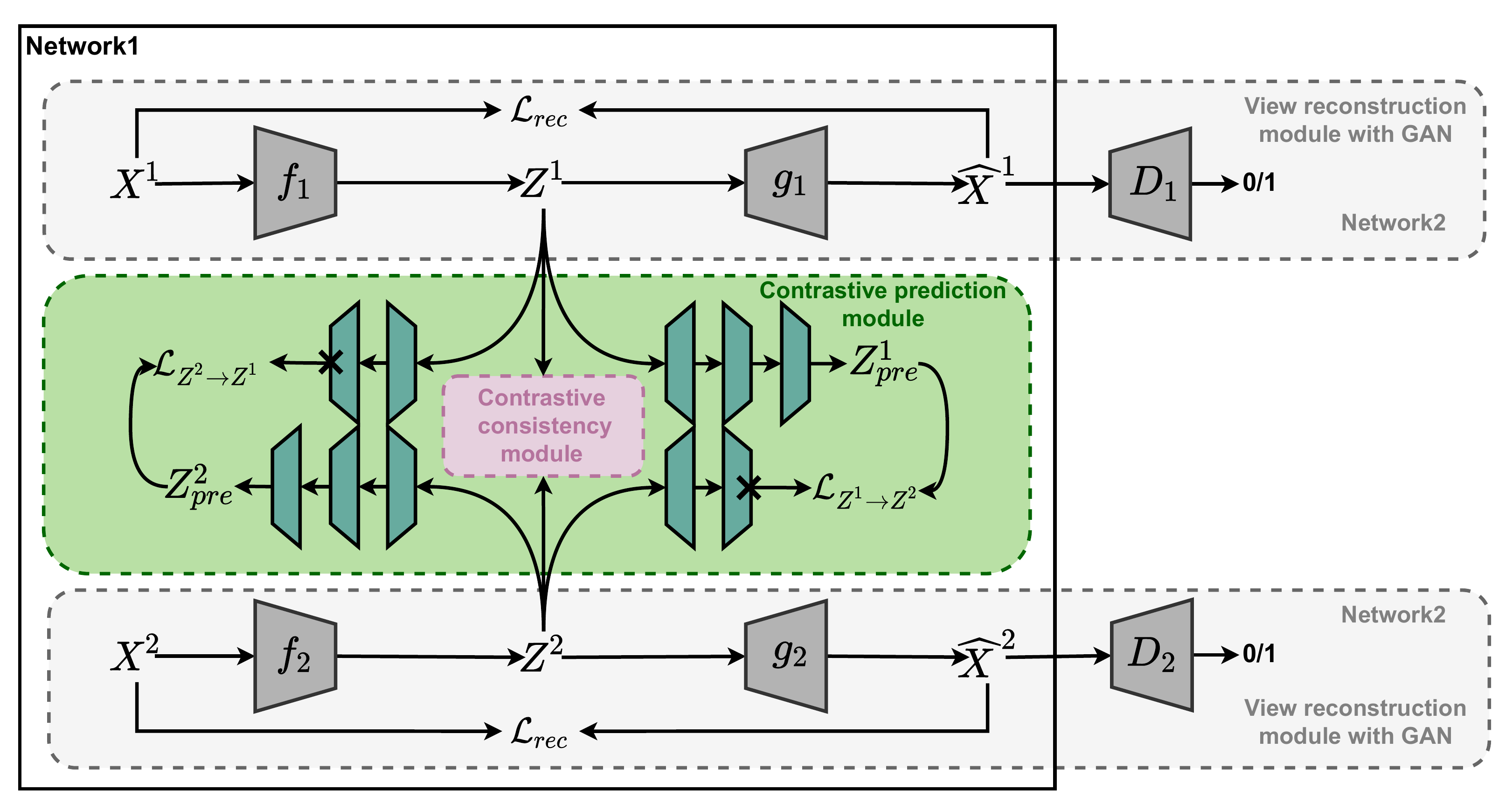}
  \caption{The framework of the proposed CIMIC-GAN. CIMIC-GAN consists of network1 which processes complete data and Network2 which processes incomplete data.  \textcolor{black}{To be specific, the encoder $f_{(v)}$ and decoder $g_{(v)}$ learn the latent representation $Z^{v}$ by minimizing the reconstruction loss $\mathcal{L}_{rec}$ of $X^{v}$ and $\widehat{X}^{v}$ (Section 3.3.1).} Green and pink boxes represent double contrastive learning modules in Network1.  \textcolor{black}{The contrastive prediction module trains a pair of prediction networks that make $Z^{1}$ and $Z^{2}$ predict each other (Section 3.3.3), where $Z_{pre}^{1}$ represents the prediction of $Z^{1}$ to $Z^{2}$, and $Z_{pre}^{1}$ will approximate $Z^{2}$ as much as possible. The contrastive consistency module aims to maximize the mutual information between $Z^{1}$ and $Z^{2}$ (Section 3.3.4).} In Network2, the view reconstruction module with GAN is able to restore incomplete data and use the restored data to train Network1 (Section 3.3.2).
  }\label{fig1}
  \end{center}
\end{figure*}
\begin{itemize}

    \item We address the reason why it is difficult to learn more advanced representation in IMC of images, and find it is primarily due to very few complete data pairs in practice while neglecting hidden information of incomplete data. 
    \item We incorporate GAN into feature representation to comprehensively utilize incomplete view data, which makes up for the difficulty of consistent learning when complete data is insufficient. Our proposed method can perform well even with abnormally high missing rates.
    \item The proposed auto-encoding representation maintains more complementary information in the multiple-view feature space, and benefits the overall double contrastive learning process. Ultimately, the proposed self-supervised clustering model alleviates representation inconsistency of multiple views and thus enhances consistency of self-supervised IMC.
    \item Experiments on \textcolor{black}{four} types of datasets show that CIMIC-GAN achieves state-of-the-art clustering performance and strong robustness. 
\end{itemize}

To our knowledge, CIMIC-GAN is the first clustering solution that can mine vision features both in complete views and incomplete views. Without loss of generality, we implement CIMIC-GAN in a two-view clustering scenario, but there remains space for expansion in the case of more views. This work allows IMC models to be embedded into the physical world to learn more consistent representation \textcolor{black}{in broad scenarios} in a self-supervised way.

%% file: Related.tex
\section{RELATED  WORK}\label{sec2}

In the section, we briefly review \textcolor{black}{three} lines of related work, contrastive learning ,incomplete multi-view clustering, and \textcolor{black}{generative adversarial networks}.

%\vspace{-\topsep}
\subsection {Contrastive Learning}

Contrastive learning \cite{24,25,26,27} is an essential method for unsupervised learning \cite{28}. Its major goal is to maximize feature space similarity between positive samples while reducing the distance between negative samples. Most contrastive learning methods \cite{36,37} are built to handle various data augmentations or modal augmentations. Especially in the field of computer vision, contrastive learning methods have produced excellent results~\cite{29}. As an example, methods such as SimClR or MoCo \cite{30,37,32} minimize the InfoNCE loss function \cite{33} to maximize the upper bound of mutual information. Because dealing with negative samples is inconvenient, later contrastive learning algorithms \cite{34,35} have successfully replaced the contrast task with the prediction task without the need for negative samples. \textcolor{black}{More tools can be used in contrastive learning. As the first comparative hashing method, CMH \cite{71} solves the problem of the negative impact of false negative samples.}

Almost all existing contrastive learning methods \cite{30,37,32,34,35} are designed to handle single-view data, exhaustively exploring various data augmentations to build different views/augments. To the best of our knowledge, this is the first time double contrastive learning has been used on the IMC problem to handle the challenge of mutual promotion of consistency learning and data recovery from a different perspective. The introduced double contrastive learning enhances consistency in learning while alleviating inconsistency. We use latent representations to learn the consistency between distinct views and a BYOL \cite{34} based prediction network to attenuate the inconsistency between different views. 
These two are integrated and mutually promote each other, which greatly improves the clustering performance.
As far as we know, this is the first work that uses the BYOL structure for data recovery.

\subsection {Incomplete Multi-view Clustering (IMC)}

\textcolor{black}{According to the utilization method of multimodal data, traditional IMC methods can be roughly divided into four categories: IMC based on kernel learning \cite{47,60}, IMC based on matrix factorization, IMC based on graph, and IMC based on spectral clustering. Trivedi et al. proposed a kernel-based method \cite{60} that recovers the kernel matrix of incomplete views from the kernel matrix of complete views but requires one of the views to be all complete. To address this limitation, matrix factorization-based IMC methods \cite{3, 4, 5, 46, 49} use lower ranks to project parts of the data into a common subspace, formally similar to the K-means relaxation method \cite{12}. In the follow-up work, Linlin et al. \cite{61} introduced an indicator matrix to map non-existing instances into 0-element vectors, and Zhiqi et al. \cite{63} introduced a fusion matrix while learning the self-representation in the subspace, which further developed the IMC method based on matrix decomposition. However, feature-based matrix factorization methods generally cannot explore nonlinear data structures, while graph-based methods \cite{13, 14} can effectively deal with nonlinear data structures. For example, PSIMVC-PG \cite{64} proposes a parameter-free clustering based on a prototype graph framework to handle non-linear data structures with lower complexity. In addition, the spectral clustering-based method \cite{52} can also deal with the nonlinearity in the data by constructing a Laplacian graph. For example, FMDC \cite{65} uses a very short time to construct an anchor graph on different views, which significantly reduces the time cost. IMSC-AGL\cite{62} used graph learning and spectral clustering to learn incomplete public representations for the first time, and achieved better results. }

\textcolor{black}{However, traditional IMC methods have limitations in terms of representation ability and high complexity. In recent years, the self-supervised IMC method \cite{6,7,8,21,44,48,51,66} based on deep learning has attracted attention, and this method combined with deep neural networks shows strong generalization ability. The differences between this work and existing studies are as follows. CIMIC-GAN belongs to a deep IMC framework, which has better scalability than traditional IMC methods. Most existing IMC methods \cite{3,4,5,8,21,45,51,52} can only infer and generate missing data with consistency learned from complete data, but can't properly use incomplete data, wasting the hidden information contained in the training set's incomplete data. In contrast, we leverage GAN to fill in the missing views and train the prediction network with the new data generated. Therefore, we target the problem that incomplete data in IMC cannot be used or trained. It can even outperform other methods in extreme circumstances where the missing rate is exceedingly high.}

\subsection {\textcolor{black}{Generative Adversarial Networks (GAN)}}

\textcolor{black}{Recently, there have been several approaches utilizing GANs \cite{40} for multi-view learning, such as CycleGAN \cite{2, 76} and StarGAN \cite{75}. Wang et al. proposed a generative adversarial network (GAN) based deep IMC method \cite{7, 40}. It uses one view to generate missing data for another view of recurrent GAN \cite{2}. However, the accompanying challenge of cross-modal generation is how to narrow the differences between different modalities. To overcome this problem, CAD \cite{72} fills the heterogeneity gap between modalities by matching the synthesized data in the feature space with the help of two groups of specific GANs and a discriminative mechanism across modalities. However, GANs mostly adopt shallow models, which are difficult to capture deep semantic understanding. AIMC \cite{8} uses deeper networks to perform missing data inference while finding the common latent space of multiple view data.}

\textcolor{black}{Different from the above works, these methods mainly focus on cross-domain data generation. In our framework, the adversarial process only acts on a single view, which preserves the private information of the view to the greatest extent, thus effectively realizing incomplete data recovery in independent feature space. Moreover, the adversarial process is integrated with the target reconstruction, so that the pseudo-data distribution contains more complementary information, which is beneficial for the network to capture a better clustering structure. Furthermore, Consistent GAN \cite{7} for the two-view IMC problem can only handle dual-view data, it uses one view to generate missing data from the other and then performs clustering on the generated full data. While CIMIC-GAN can be extended and applied to the IMC problem with an arbitrary number of views.}

%% file: Method.tex
\section{The Method}\label{sec3}

In order to better solve the IMC problem by learning consistency of incomplete multiple views, we propose Contrastive Incomplete Multi-View Image Clustering with Generative Adversarial Network (CIMIC-GAN).

\subsection{Notations and Problem Formulation}
\begin{figure*}[t]
  \begin{center}
  \includegraphics[width=0.95\textwidth]{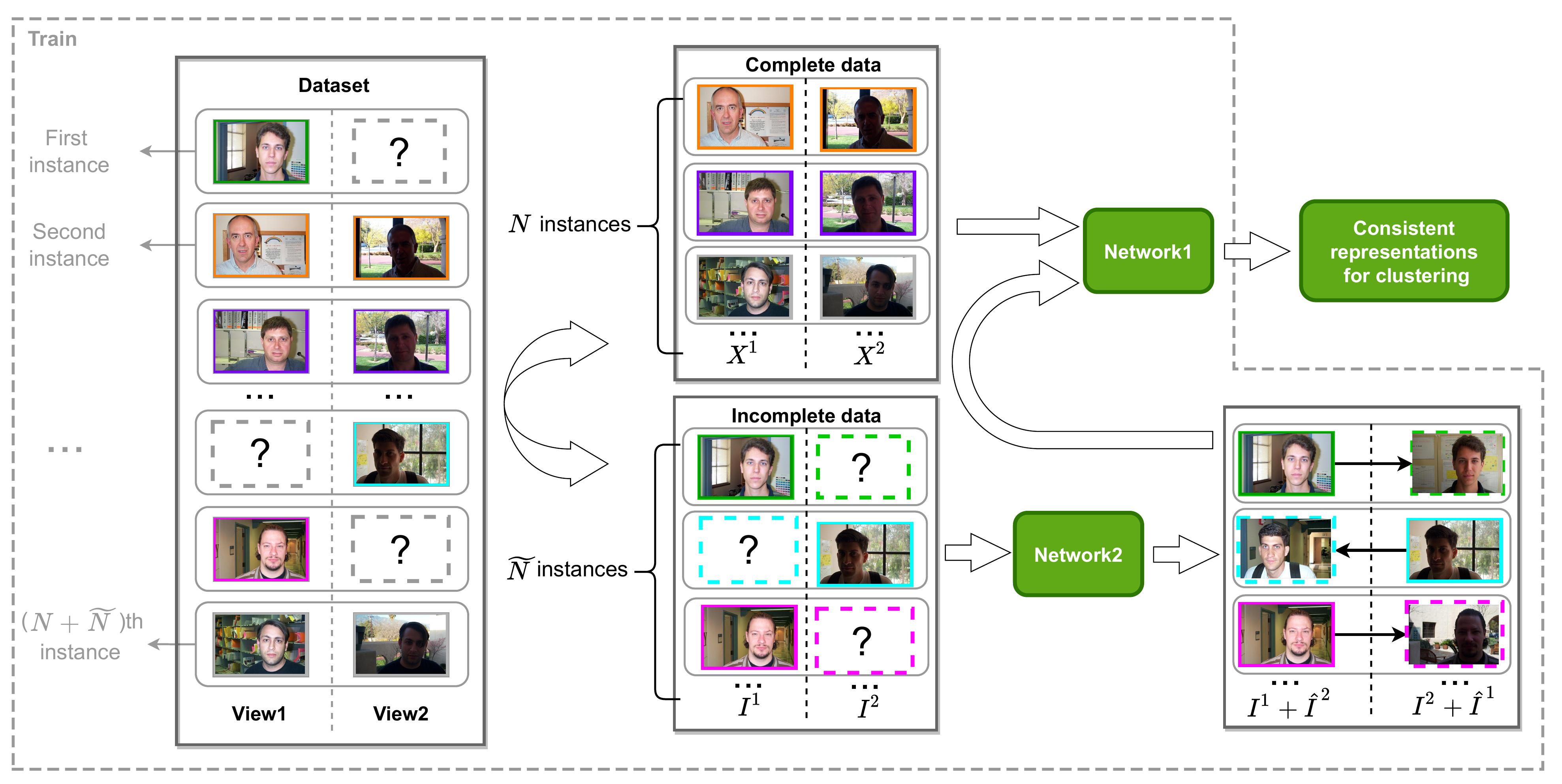}
  \caption{
 Data flow and the training process. The multi-view data can be divided into two parts: paired complete data and incomplete data. Incomplete data can be restored and completed through Network2. Then the restored data along with the original complete data can be used as the input of Network1 for consistency representation learning and K-means clustering.
  }\label{fig2}
  \end{center}
\end{figure*}
Given an instance, it usually contains multiple views or multiple modalities, such as RGB images or depth maps, representing different perspectives of the same object. Regarding this phenomenon, the incomplete multi-view clustering problem tries to solve the premise that there are missing views in an instance and can have good clustering performance. 
One instance generally has multiple views, which may or may not be complete. As shown in Fig. \ref{fig2}, assuming that a dataset has two views, namely $A$=2, there are $(N+\widetilde{N})$ instances in total, such a dataset can be divided into two parts: complete data and incomplete data. In the set of complete instances, we use $X_{n}^{(v)}(v=1, \ldots, A)$ to denote the feature vector for the $v$-th view of the $n$-th instance. In the set of incomplete instances, we use $I_{\widetilde{n}}^{(v)}(v=1, \ldots, A)$ to denote the feature vector for the $v$-th view of the $\widetilde{n}$-th instance. $\forall X_{n}^{(v)}, I_{\tilde{n}}^{(v)} \in \mathbb{R}^{d_{v}}$ , where $d_{v}$ is the dimensionality of the $v$-th view. Our goal is to group all the $N+\widetilde{N}$ instances into $K$ clusters.

Without loss of generality, we make $A=2$, \emph{i.e.}, two views are involved. Define a two-view dataset as a collection of $\left\{{X}^{1}, {X}^{2},{I}^{1},{I}^{2}\right\}$ of $(N+\widetilde{N})$ instances, where $X^{1}$ and $X^{2}$ represent instances that exist and are aligned in two views, and where $I^{1}$ and $I^{2}$ represent instances that exist only in the first and second views, respectively. 

\subsection{Model Overview}

Multi-View Clustering is an unsupervised learning problem. The key to solve this problem is to learn strong representations as well as the consistency of them. Considering that contrastive learning has a strong representation ability in the field of unsupervised learning, we incorporate a double contrastive learning module (Network1) into our model, CIMIC-GAN, to learn consistency of the representations of the views. On the other hand, considering the missing data of instances, we fill the incomplete data in a generative adversarial way to obtain more complete auto-encoding representations of views (Network2).  The training process of CIMIC-GAN is divided into three steps:

%CIMIC-GAN consists of two groups of networks, Network1 \notes{Use uppercase or lowecase for both the figure and here. Stay consistent.(Wang:Arleady modified)} for processing aligned complete data $\left\{X^{1}, X^{2}\right\}$ and Network2 for processing misaligned incomplete view data $\left\{I^{1}, I^{2}\right\}$. Network1 is composed of an autoencoder and a mutual prediction network. On the other hand, Network2 adds a decoder to each autoencoder to form a generative adversarial network (GAN) for missing data generation of the incomplete view. The CIMIC-GAN training process is divided into three steps:

\emph{Step 1: Train Network1 with $\left\{X^{1}, X^{2}\right\}$.} The aligned complete view data $X^{1}$ and $X^{2}$ is respectively used in the encoder $f_{1}$ and $f_{2}$ of Network1 to obtain the latent representations. $Z^{1}$ and $Z^{2}$ is the latent representations of the first view and the second view, respectively. Based on $Z^{1}$ and $Z^{2}$, there are three objective functions that need to be optimized in this step: i) The loss obtained by reconstructing different views through the autoencoder is denoted by $\mathcal{L}_{rec}$. ii) By using contrastive learning, we maximize mutual information between $Z^{1}$ and $Z^{2}$. The corresponding loss is denoted by $\mathcal{L}_{cc}$. iii) Through contrastive learning without negative samples, $Z^{1}$ and $Z^{2}$ are predicted by two symmetrical networks using predictors to alleviate the inconsistency between distinct viewpoints, and the loss function is denoted by $\mathcal{L}_{pre}$. Ultimately, we propose the overall objective function of network 1:
\begin{equation}
\mathcal{L}=\mathcal{L}_{cc}+\alpha \mathcal{L}_{rec}+\beta \mathcal{L}_{pre}.
\label{eq101}
\end{equation}
The parameters $\alpha$ and $\beta$ are the balanced factors on $\mathcal{L}_{rec}$ and $\mathcal{L}_{pre}$, respectively. We set these two parameters as 0.1, which will be proven in the following parameter analysis experiments.

\emph{Step 2: Train Network2 using $\left\{I^{1}, I^{2}\right\}$.} Feed the incomplete view data $I^{1}$ and $I^{2}$ into the encoder of Network2. Note that, the encoder $f_{1}$, $f_{2}$ and the decoder $g_{1}$, $g_{2}$ already have converged in step 1. The decoder $g_{1}$ and $g_{2}$ will be equivalent to being well initialized as generators in the GAN structure. Each decoder incorporates the corresponding discriminator $D_{(v)}$ to form a typical GAN network. In detail, $\widehat{I}^{(v)}$ is first generated according to $I^{(v)}$. Then, the discriminator $D_{(v)}$ will judge whether $\widehat{I}^{(v)}$ generated by the decoder is true. Till the discriminator cannot provide that correctly, the generator $g_{(v)}$ converges. The purpose of this step is to train a powerful generator to generate the missing data of imcomplete view as well as expand the training dataset. As shown in Figure 2, according to the incomplete view $I^{(v)}$, the corresponding missing data $\widehat{I}^{(v)}$ is generated through Network2 and fills it in the corresponding modality in pair,  $\left\{I^{1}+\widehat{I}^{2}, I^{2}+\widehat{I}^{1}\right\}$.

\emph{Step 3: Train Network1 again with $\left\{I^{1}+\widehat{I}^{2} , I^{2}+\widehat{I}^{1}\right\}$ to achieve a consistent representation. }
Feed the pseudo-complete data of different views $\left\{I^{1}+\widehat{I}^{2}, I^{2}+\widehat{I}^{1}\right\}$ to Network1, and perform consistency learning as step 1. The optimization objective is the same as the first step, and the balance factor has not changed, as follow Eq.(\ref{eq101}). The role of this step is to get sufficient training data to make the Network1 model more generalized and robust.
After Network1 has converged, we feed $\left\{{X}^{1}, {X}^{2},{I}^{1},{I}^{2}\right\}$ to Network1. This will give us latent representations of all views, including views with missing data. Finally, a consistent representation is obtained by concatenating representations of different views (see \ref{fig2}), which is further input to a specific cluster (\textcolor{black}{Kmeans or other adaptive deep clustering methods with enhanced performance, such as GATCluster \cite{67}, Spice\cite{68}, etc.}) to achieve qualified multi-view image clustering.

\subsection{Loss Function}
The loss functions used in the proposed model is provided in this section.

\subsubsection{Reconstruction Loss}

\begin{figure}[h]
\centering
\includegraphics[width=0.9\linewidth]{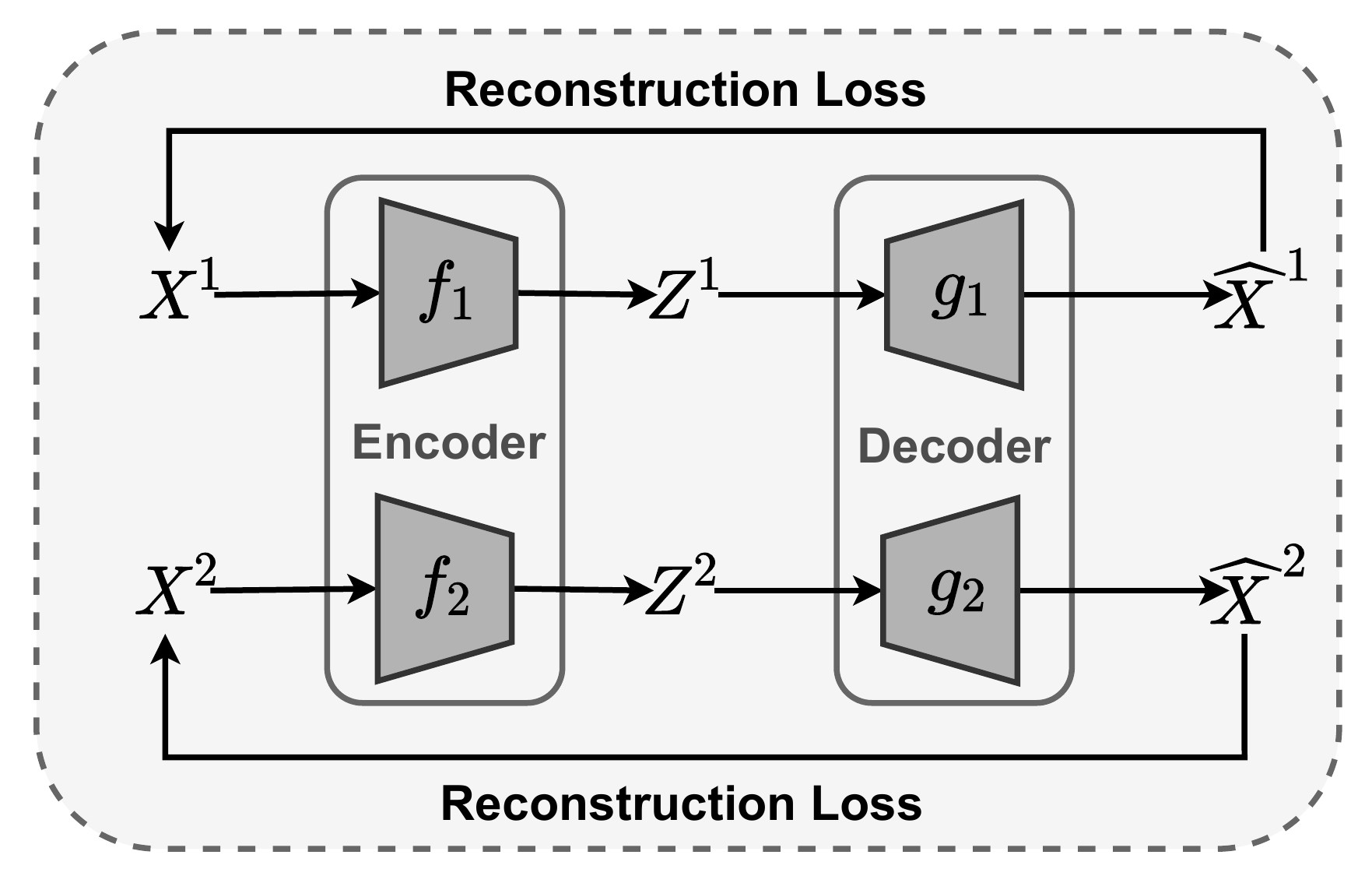}
\caption{View reconstruction module}
\label{fig3}
\end{figure}

Autoencoders \cite{38} can compute nonlinear mappings by learning the latent feature space between input and output. As shown in Fig. \ref{fig3}, we designed a deep autoencoder to project a view into its latent space while learning latent representation. The structure of this autoencoder ensures a diversity of viewpoints while mining complementary information from more clustering patterns, which is the foundation to improve clustering performance with multimodal data. Additionally, avoiding model collapse is advantageous. 
$\hat{X}^{(v)}$ is $X^{(v)}$ the result of  being reconstructed by the autoencoder. We use $L_{2}$ distance as our view reconstruction loss function , $\mathcal{L}_{rec}$ is defined as
\begin{equation}
\mathcal{L}_{r e c}=\sum_{v=1}^{2} \sum_{n=1}^{N}\left\|X_{n}^{(v)}-g_{(v)}\left(Z_{n}^{v}\right)\right\|_{2}^{2},
\label{eq2}
\end{equation}
where $X^{(v)}_{n}$ denotes the $n$-th sample of $X^{v}$. $g_{(v)}$ denotes the decoder for the $v$-th view. The latent representation of $n$-th sample in $v$-th view is given by
\begin{equation}
Z_{n}^{v}= f_{(v)}\left(X_{n}^{(v)}\right),
\label{eq3}
\end{equation}
where $Z^{(v)}$ denotes the representations of $X^{(v)}$ and $f_{(v)}$ denotes the encoder for the $v$-th view,$v \in\{1,2\}$.

\subsubsection{Adversarial Loss}

\begin{figure}[h]
\centering
\includegraphics[width=0.9\linewidth]{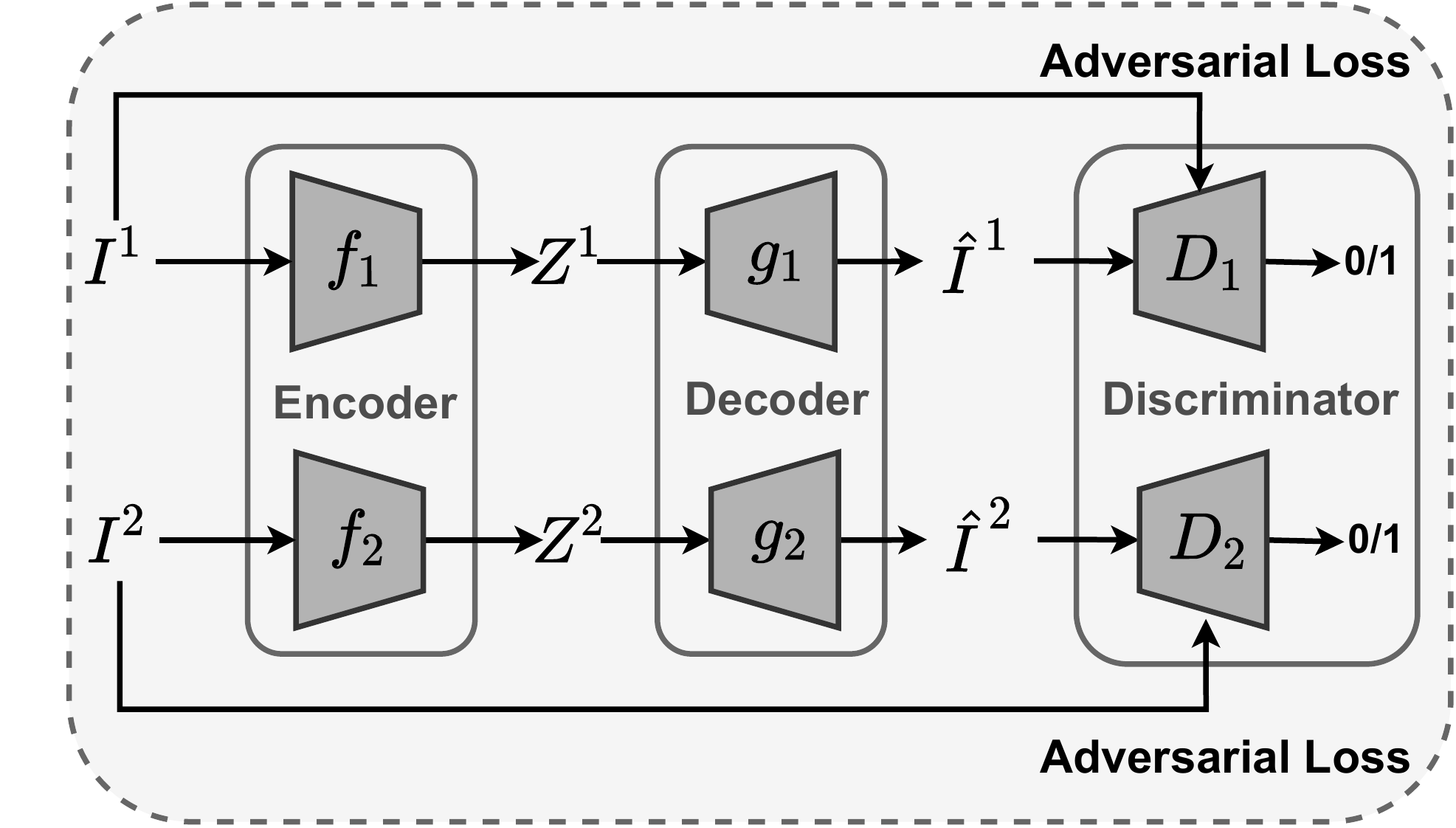}
\caption{View reconstruction module with GAN.}
\label{fig5}
\end{figure}

As shown in Fig. \ref{fig5}, the decoder $g_{(v)}$ and the discriminator $D_{(v)}$ constitute two sets of GANs. $D_{(v)}$ essentially matches the difference between the latent representation distribution and the data distribution, with the goal of aligning the different representation distributions. In step2 (see Section 3.2), the corresponding data in the missing view is generated according to the incomplete data, and combined into a new data set $\left\{I^{1}+\widehat{I}^{2}, I^{2}+\widehat{I}^{1}\right\}$.

Adversarial loss is based on a paper \cite{40}. The traditional GAN will train two models of generator $G$ and discriminator $D$, and the objective function is the value function defined as:
\begin{equation}
\begin{split}
\min _{G} \max _{D} V(D, G)=\mathbb{E}_{\boldsymbol{x} \sim p_{\text {data }}(\boldsymbol{x})}[\log D(\boldsymbol{x})]+\\
\mathbb{E}_{\boldsymbol{z} \sim p_{\boldsymbol{z}}(\boldsymbol{z})}[\log (1-D(G(\boldsymbol{z})))].
\label{eq12}
\end{split}
\end{equation}

$V(D,G)$ is a two-player minimax game with two expectations added. Leveraging the decoder as a generator, we implement a generative adversarial framework for multi-view reconstruction, and thus the corresponding loss function is designed as:
% \begin{flalign}\label{eq12}
% \mathcal{L}_{adv}=\min _{\left\{g_{v}\right\}_{v=1}^{A}} \underset{\left\{D_{v}\right\}_{v=1}^{A}}{\max }
% &\sum_{v=1}^{A}\left(\mathbb{E}_{\mathbf{x}^{(v)} \sim P_{I^{(v)}}}\left[\log D_{v}\left(I^{(v)}\right)\right]\right.\left.\\
% &+\mathbb{E}_{\mathbf{z} \sim P_{Z^{v}}}\left[\log \left(1-D_{v}\left(g_{v}(\mathbf{z})\right)\right)\right]\right).
% \end{flalign}
\begin{equation}
\label{eq12}
\begin{aligned}
\mathcal{L}_{adv}=\min_{\{g_{v}\}_{v=1}^{A}} \underset{\{D_{v}\}_{v=1}^{A}}{\max }
&\sum_{v=1}^{A}(\mathbb{E}_{\mathbf{x}^{(v)} \sim P_{I^{(v)}}}[\log D_{v}(I^{(v)})]\\
&+\mathbb{E}_{\mathbf{z} \sim P_{Z^{v}}}[\log (1-D_{v}(g_{v}(\mathbf{z})))])
\end{aligned}
\end{equation}
we reconstruct the adversarial loss of incomplete data $I^{v}$ according to a view-specific latent representation $Z^{(v)}$, and the representation of the missing view $\widehat{I}^{(v)}$ is obtained.

The missing data representation generated by $g_{(v)}$ incorporates $\left\{I^{(v)}\right\}$ to form the representation of a complete view, $\left\{I^{1}+\widehat{I}^{2}, I^{2}+\widehat{I}^{1}\right\}$, which is used to retrain Network1 in step3 (see Section 3.2). This achieves more effective use of incomplete data $\left\{I_{\widetilde{n}}^{(v)}\right\}$, and makes Network1 more fully trained.

\subsubsection{Contrastive Prediction Loss}

\begin{figure*}[t]
  \begin{center}
  \includegraphics[width=0.85\textwidth]{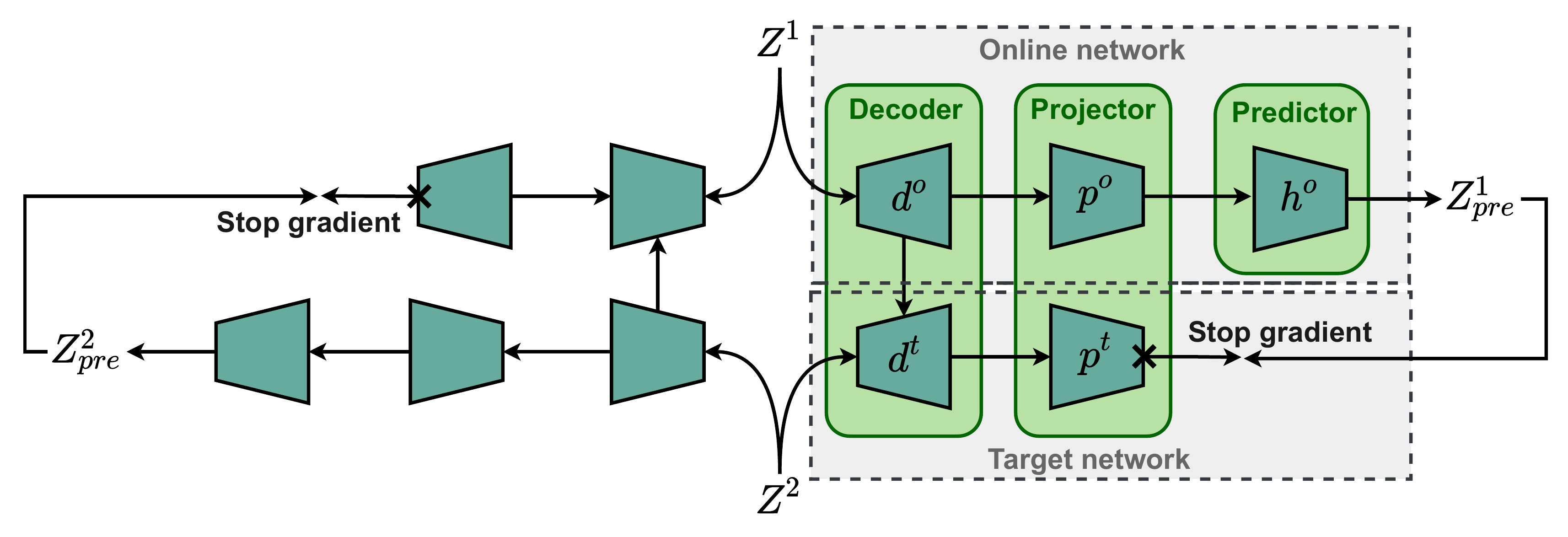}
  \caption{Illustration of the contrastive prediction module. All the projectors and predictors in this figure are MLPs. Since this module is left-right symmetrical, we take the right part as an example.}\label{fig4}
  \end{center}
\end{figure*}
We propose a mutual prediction methods based on contrastive learning without negative samples to address the problem of missing data recovery. 
In contrast to the traditional contrastive learning  \cite{34}, every decoder and Multi-Layer Perceptron (MLP) in the contrastive prediction module take a softmax function to achieve explicit prediction. Additionally, we design two networks with the same structure, where each predictor can predict the latent representation of the corresponding modality.

As shown in Fig. \ref{fig4}, the online network consists of a decoder $d^{o}$, a projector $p^{o}$, and another predictor $h^{o}$. On the other hand, the target network includes a decoder $d^{t}$ and a decoder $p^{t}$. With using a mean squared error loss function, the outputs of the online network and the target network approach consistent.
This process can be formulated by the following loss function:

\begin{align}
\mathcal{L}_{Z^{1} \rightarrow Z^{2}} &=\left\|Z^{1}_{pre}-p^{t}\left(d^{t}\left(Z^{2}\right)\right)\right\|_{2}^{2}\nonumber\\
&=2-2 \cdot \frac{\left\langle Z^{1}_{pre},p^{t}\left(d^{t}\left(Z^{2}\right)\right)\right\rangle}{\left\|Z^{1}_{pre}\right\|_{2} \cdot\left\|{p^{t}\left(d^{t}\left(Z^{2}\right)\right)}\right\|_{2}},
\label{eq7}
\end{align}

where the output obtained by the latent representation $Z^{1}$ through the online network is defined by
\begin{equation}
Z^{1}_{pre} =h^{o}\left(p^{o}\left(d^{o}\left(Z^{1}\right)\right)\right).
\label{eq8}
\end{equation}
After that, %$Z^{1}$ and $Z^{2}$ will also perform a reversal, 
we send $Z^{1}$ to the target network, and send $Z^{2}$ to the online network. This process is defined as
\begin{equation}
\mathcal{L}_{Z^{2} \rightarrow Z^{1}} =\left\|h^{o}\left(p^{o}\left(d^{o}\left(Z^{2}\right)\right)\right)-p^{t}\left(d^{t}\left(Z^{1}\right)\right)\right\|_{2}^{2}.
\label{eq9}
\end{equation}
The loss function of the whole  can be defined as
\begin{equation}
\mathcal{L}_{pre}=\mathcal{L}_{Z^{1} \rightarrow Z^{2}}+\mathcal{L}_{Z^{2} \rightarrow Z^{1}},
\label{eq10}
\end{equation}
because the contrastive prediction module is two sets of left and right symmetrical modules, so it is still necessary to add the loss of another set of prediction network.

Decoder $d^{o}$ and decoder $d^{t}$ use the same network architecture but different parameters, decoder $d^{o}$ is updated with the gradient update, and decoder $d^{t}$ is updated in the form of moving average like \cite{37}, that is, a momentum encoder is used.
Decoder $d^{t}$ provides the regression target while training decoder $d^{o}$, and its parameters $\omega^{t}$ are an exponential moving average of decoder $d^{o}$ ’s parameters $\omega^{o}$. To update these parameters, a target momentum $m \in [0,1]$ is involved as the following,

\begin{equation}
\omega^{t} \leftarrow m \omega^{t}+(1-m) \omega^{o}
\label{eq1}
\end{equation}
If we choose a large momentum during training, the decoder $d^{t}$ updates very slowly and does not change rapidly with decoder $d^{o}$, thus ensuring that decoder $d^{t}$ is always a similar encoder at each step.

\subsubsection{Contrastive Consistency Loss}

\begin{figure}[h]
\centering
\includegraphics[width=0.6\linewidth]{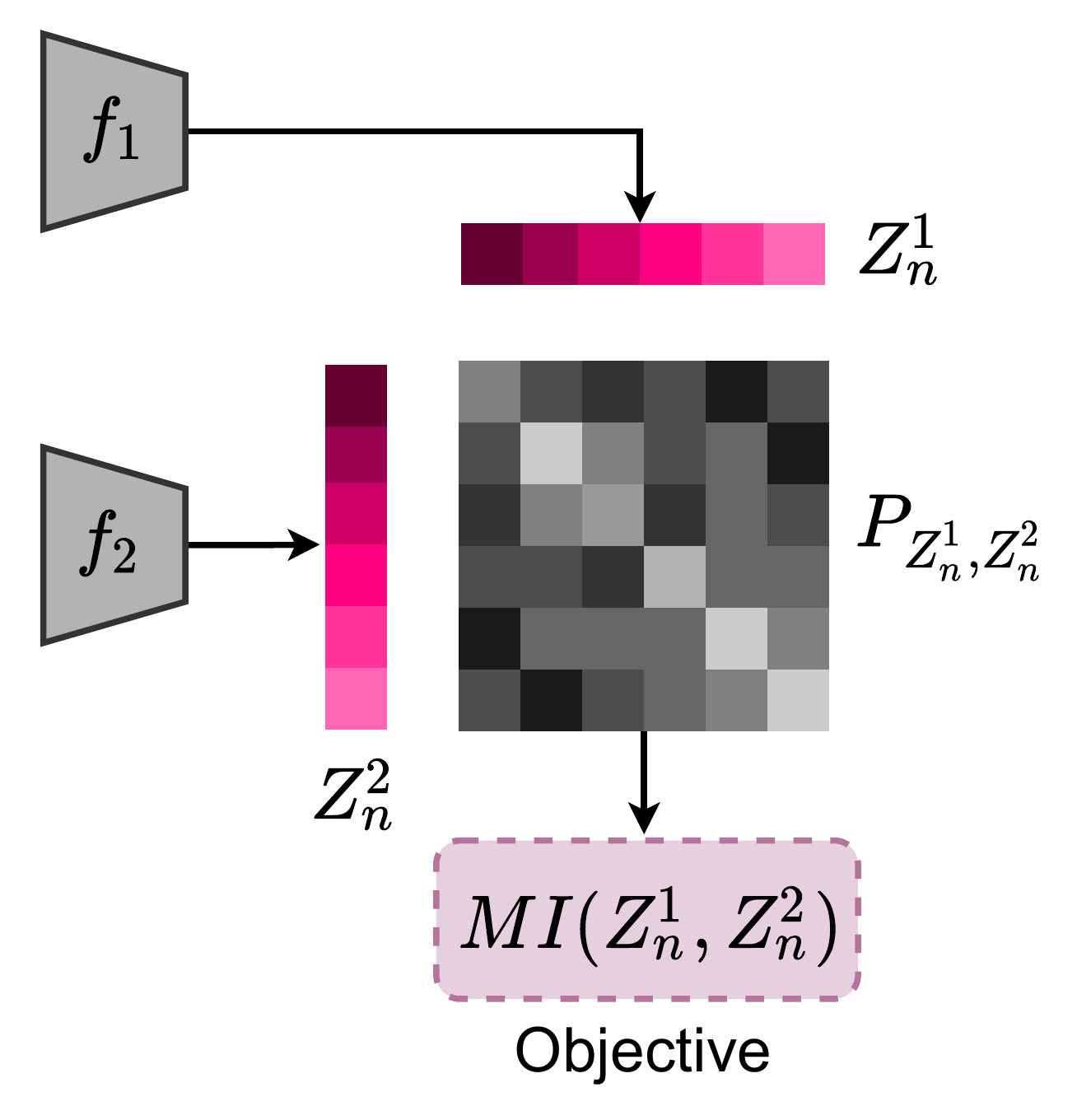}
\caption{\textcolor{black}{Contrastive consistent module. The mutual information of $Z^{1}_{n}$ and $Z^{2}_{n}$ can be directly obtained on a joint probability distribution matrix $P_{Z_n^1,Z_n^2}$. The matrix can be calculated by approximating $Z^{1}_{n}$ and $Z^{2}_{n}$ as two independent discrete probability distributions. %The marginal probability distributions $P_{Z_n^1}$ and$P_{Z_n^2}$ can be obtained by summing the rows and columns of $P_{Z_n^1,Z_n^2}$, and then the mutual information can be directly obtained according to Formula \ref{eq12}.
}
}
\label{fig334}
\end{figure}

Inter-view consistency learning and missing view completion are mutually reinforcing processes~\cite{21}. A common strategy to maximize the consistency between two views is to maximize the correlation, which can also be understood as extracting shared components. Here, maximizing the consistency between $Z^{1}$ and $Z^{2}$ can be formulated as maximizing the mutual information of $Z^{1}$ and $Z^{2}$, as shown in Fig. \ref{fig334}.

\begin{equation}
{{\mathcal{L}}_{cc}}=-\sum\limits_{n=1}^{N}{\left( MI\left( Z_{n}^{1},Z_{n}^{2} \right)+\gamma \left( H\left( Z_{n}^{1} \right)+H\left( Z_{n}^{2} \right) \right) \right)},
\label{eq4444}
\end{equation}
where $H$ is information entropy, $MI$ is mutual information, and a regularization term~\cite{21} with  parameter $\gamma = 9$ is involved to improve the generalization of the module. In order to calculate the mutual information $MI\left(Z_{n}^{1}, Z_{n}^{2}\right)$, the output of the softmax function in the decoder involved in Section 3.3.3 is used as an over-cluster class probability distribution of $Z_{n}^{*}$ \cite{56}, and the joint probability distribution of  $Z_{n}^{1}$ and  $Z_{n}^{2}$ is obtained. In this way, $MI$ is calculated according the following formula~\cite{21}, 

\begin{equation}
\mathbb{E}_{P_{Z_n^1,Z_n^2}} \left( P_{Z_n^1,Z_n^2}log\frac{P_{Z_n^1,Z_n^2}}{{P_{Z_n^1}^{\gamma+1}}P_{Z_n^2}^{\gamma+1}}\right)
\label{eq12}
\end{equation}
where $\gamma$ is the balance term defined in Formula \ref{eq4444}.

%% file: Experiment.tex
\section{EXPERIMENTS}\label{sec4}

In this section, we evaluate the proposed CIMIC-GAN method on \textcolor{black}{four} widely-used multi-view datasets and compare it with several state-of-the-art clustering methods.
\subsection{Datasets and Experimental Settings}
We evaluate CIMIC-GAN on \textcolor{black}{four} clustering problem datasets. 1) Caltech101-20  \cite{41} consists of 2,386 images of 20 subjects, and we use two views of HOG and GIST features, with 1984 and 512 as the feature dimensions, respectively. 2) Scene-15 \cite{42} consists of 4,485 images distributed over 15 Scene categories, and we use two views of PHOG and GIST features, 20D and 59D feature vectors, respectively. 3) LandUse-21 \cite{43} consists of 2100 satellite images from 21 categories , and we use two views of PHOG and LBP features, 59D and 40D feature vectors, respectively. 4) \textcolor{black}{A large dataset, Noisy MNIST \cite{48}, consists of 70,000 instances of 10 classes. We randomly select 15,000 original instances as view 1 and 15,000 Gaussian noise-added instances as view 2.} We summarize the detailed statistics of the datasets in Table \ref{table1}. %\revs{CIMIC-GAN mainly focuses on two-view clustering, and there is still room for expansion in the case of more views.} \notes{Can we use more than two views, like 3 views? It would be better if we use like 3 views in one of the three datasets. Wang:Already modified}
\begin{table}[H]
\renewcommand\arraystretch{1.3}
\centering
\caption{Dataset summary}
\label{table1}
\setlength{\tabcolsep}{3mm}{
\begin{tabular}{cccc}

\toprule \text Datasets     & Size      & \multicolumn{1}{l}{\# of categories} & Dimension     \\
\hline \text Caltech101-20      & 2386      & 20                                  & 1984/512 \\
             Scene-15        & 4485      & 15                                  & 20/59  \\
             LandUse-21      &2100       &21                                   & 59/40 \\
             20     &70,000  &10  &784/784  \\
\bottomrule  \text

\end{tabular}
}
\end{table}

We conduct all the experiments on the platform of ubuntu 16.04 with Tesla P100 Graphics Processing Units (GPUs) and 32G memory size. Our model, method and baseline are built on the pytorch 1.7.0 framework. Based on extensive ablation studies, the bacth size is set to 256 and the epochs of the three steps of training are 300, 250, and 200, respectively. The missing rate is fixed to 0.5, the momentum $m$ is fixed to 0.6, the entropy parameter $\gamma$ is fixed to 9 and trade-off hyper-parameters $\alpha$ and $\beta$ are fixed to 0.1. We utilize Adam optimizer \cite{54} with default parameters and a learning rate of 0.0001. We set the dimension of the autoencoder to d-1024-1024-1024-128, where d is the dimension of the input data. We simply adopt a dense (i.e., fully-connected) network where each layer is followed by a batch normalization layer and a ReLU layer. The structures of the autoencoder for different modalities are the same. MLPs are used to implement the contrastive prediction module, and all MLPs use batch normalization after each linear layer. Each MLP has two linear layers, with the ReLU activation function added in the middle of each of them.

\textbf{Compared methods and evaluate metrics:} We compare CIMIC-GAN with several baselines, including Autoencoder in Autoencoder Networks (AE2Nets)  \cite{44}, Unified Embedding Alignment Framework (UEAF) \cite{45}, Doubly Aligned Incomplete Multi-view Clustering (DAIMC) \cite{46}, Efficient and Effective Regularized Incomplete Multi-view Clustering(EERIMVC) \cite{47}, Deep Canonically Correlated Autoencoders (DCCAE) \cite{48}, Partial Multi-View Clustering (PVC) \cite{49}, Binary Multi-view Clustering (BMVC) \cite{50}, Deep Canonically Correlated Analysis (DCCA) \cite{51}, Perturbation oriented Incomplete Multi-view Clustering (PIC) \cite{52}, and COMPLETER ~\cite{21}. The clustering performance is evaluated  with three metrics: Accuracy (ACC), Normalized Mutual Information (NMI) and Adjusted Rand Index (ARI). More details on these indicators can be found in \cite{53}. A higher value of these evaluation indicators can reflect an advanced clustering performance. 

\textbf{Missing rate:} To uniformly evaluate the performance of CIMIC-GAN on incomplete multi-view data, we randomly select $\widetilde{N}$ instances as incomplete data and randomly remove some views from each of them. The Missing Rate(MR) is defined as $\frac{\widetilde{N}}{N+\widetilde{N}}$. The larger the missing rate, the more incomplete data.

\begin{figure}[h]
\centering
\includegraphics[width=0.95\linewidth]{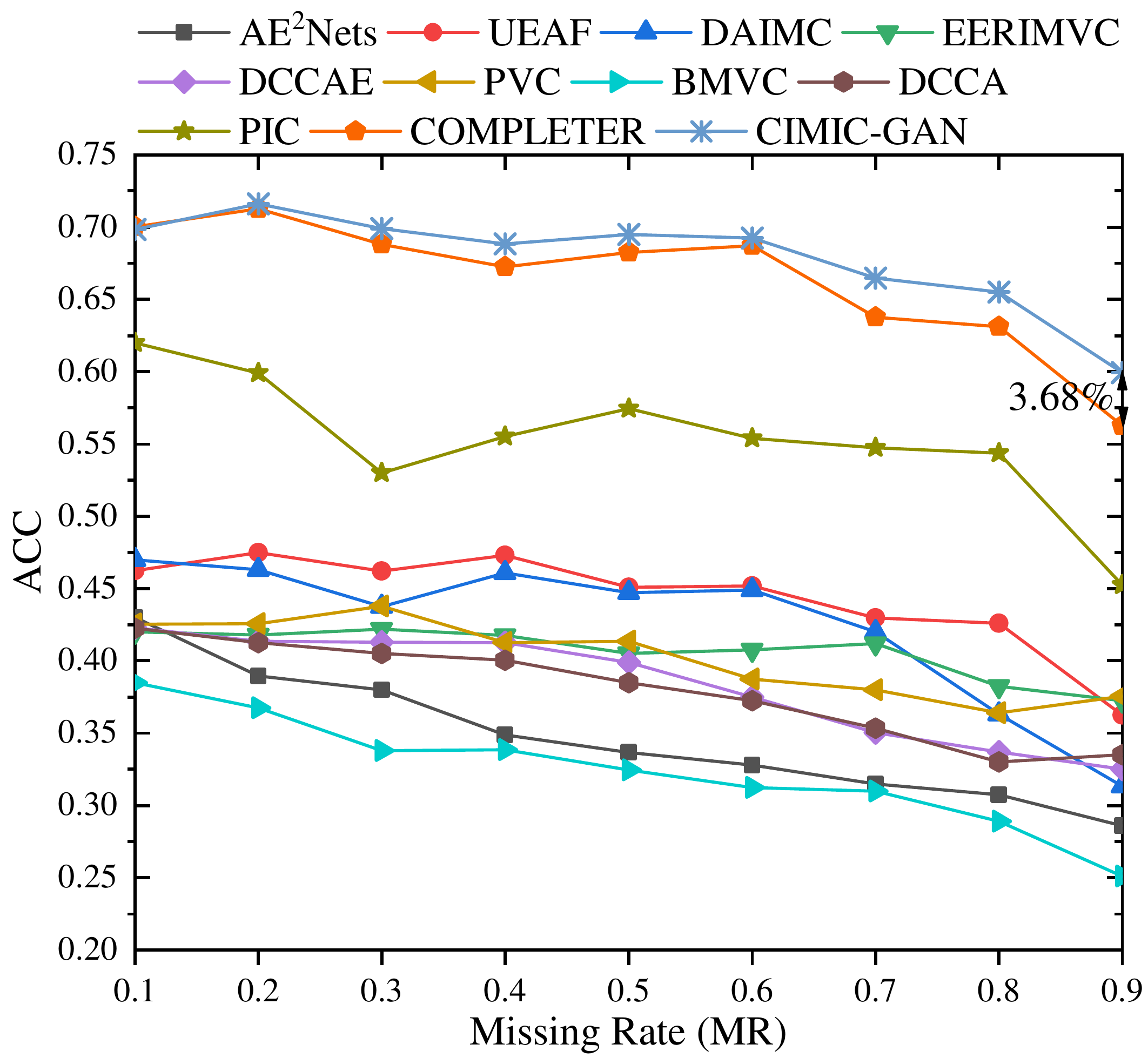}
\caption{\textcolor{black}{ACC on Caltech101-20 with different missing rates (MR).} }
\label{fig6}
\end{figure}

\subsection {Performance with Different Missing Rates}
We design experiments with different methods at different missing rates to further demonstrate the efficiency of our method at high missing rates. On the Caltech101-20 dataset, we divide the missing rate into 10 cases ranging from 0 to 0.9 with a 0.1 interval. We adjust the batch size to 128 when the missing rate is 0.9. When the missing rate is 0, since there is no incomplete data, we will directly skip the first and second steps, and we use the complete data $\left\{X^{1}, X^{2}\right\}$ to train CIMIC-GAN in an end-to-end manner. 

\textcolor{black}{Both DCCA and DCCAE learn nonlinear analysis between paired views through neural networks. BMVC needs paired views to encode and embed in binary space, and AE$^{2}$Nets needs to first convert different views into a unified comprehensive representation. However, when the missing rate is 1, it means that there are no paired samples in the original data distribution, which makes the above four methods not work. COMPLETER relies heavily on the mutual information of complete sample pairs, so the phenomenon of model collapse occurs, and the accuracy rate is only 26.33\% when all views are missing. The accuracies of UEAF, DAIMC, EEIMVC, PVC, PIC, and CIMIC-GAN are 32.15\%, 26.77\%, 35.00\%, 33.24\%, 37.12\%, and 34.30\%, respectively, in terms of the different number of missing views. The two traditional methods, PIC and EERIMVC, use the similarity of other views to fill in the missing similar items. Although the filling method is simple, it is effective when the missing rate is 1. However, our method is a deep method and infers missing data rather than missing similarities, so it enjoys higher interpretability. Even without paired samples, our method still approximates the original data distribution and achieves near-optimal performance.}

As shown in Fig. \ref{fig6}, we observe the accuracy of different methods under different missing rates and can get the following results: 1. CIMIC-GAN outperforms almost all tested baselines in all missing rates, which validates the effectiveness of the proposed method. \textcolor{black}{2. Due to the pseudo-data distribution generated by GAN fitting more missing samples, CIMIC-GAN outperforms the suboptimal method by 3.68\% with a missing rate of 0.9. This experiment also demonstrates that the model has strong robustness and generalization ability.}

\begin{table*}[ht]
\caption{The performance comparisions on \textcolor{red}{four} datasets. The ${1^{\mathrm{st}}}$ best results are indicated in \color{red}red \color{black}and the ${2^{\text {nd }}}$ best results are indicated in \color{blue}blue. \color{black}"-" indicates unavailable results due to out of memory}
\label{table2}
\renewcommand\arraystretch{1.3}
\centering
\setlength{\tabcolsep}{2.3mm}{
\begin{tabular}{llcccccccccccc}   %从哪对齐
\toprule
\multirow{2}{*}{\begin{tabular}[c]{@{}c@{}}Missing\\ Rate\end{tabular}} & \multicolumn{1}{c}{\multirow{2}{*}{Method}} & \multicolumn{3}{c}{Caltech101-20} & \multicolumn{3}{c}{Scene-15} & \multicolumn{3}{c}{LandUse-21} & \multicolumn{3}{c}{\textcolor{black}{Noisy MNIST}}\\
                                                                        &  \multicolumn{1}{c}{}                        & ACC     & NMI     & ARI     & ACC     & NMI    & ARI    & ACC     & NMI     & ARI     & ACC     & NMI     & ARI\\ \hline
\multirow{11}{*}{0.5}                                                   & AE$^{2}$Nets                                      & 33.61   & 49.12   & 24.59   & 28.01   & 31.12  & 13.85  & 19.03   & 23.01   & 5.65    & 37.98   & 33.55   & 19.08\\
                                                                        & UEAF                                        & 47.26   & 56.69   & 33.61   & 23.79   & 25.33  & 8.98   & 15.36   & 22.63   & 3.46    & 34.48   & 33.13   & 23.78\\
                                                                        & DAIMC                                       & 44.50   & 59.44   & 32.67   & 23.58   & 21.81  & 9.33   & 19.02   & 19.38   & 5.62    & 34.23   & 27.13   & 16.33\\
                                                                        & EERIMVC                                     & 40.59   & 51.25   & 27.97   & 33.04   & 31.81  & 15.92  & 22.06   & 24.96   & 9.01    & 54.99   & 45.16   & 36.31\\
                                                                        & DCCAE                                       & 39.99   & 52.70   & 29.98   & 31.79   & 34.35  & 15.67  & 14.86   & 20.89   & 3.53    & 61.23   & 59.10   & 33.25\\
                                                                        & PVC                                         & 41.34   & 56.46   & 30.75   & 25.42   & 25.20  & 11.28  & 21.26   & 22.93   & 8.07    & 36.21   & 27.92   & 17.55\\
                                                                        & BMVC                                        & 32.06   & 40.58   & 12.16   & 30.82   & 30.19  & 10.80  & 18.71   & 18.68   & 3.66    & 25.12   & 15.79   & 7.09\\
                                                                        & DCCA                                        & 38.57   & 52.42   & 29.71   & 31.73   & 33.11  & 14.85  & 13.93   & 19.91   & 3.32    & 61.99   & 60.56   & 38.09\\
                                                                        & PIC                                         & 57.49   & 64.20   & 45.18   & 38.69   & 37.97  & 20.06  & \color{blue}23.52   & 25.44   & 9.38   & -   & -   & -\\
        & COMPLETER  & \color{blue}68.47   & \color{blue}67.54   & \color{blue}74.36   & \color{blue}38.70   & \color{red}48.26  & \color{blue}23.48  & 22.15   & \color{blue}27.01   & \color{blue}10.36   & \color{blue}80.02   & \color{blue}75.33  & \color{blue}70.70\\
& \textbf{CIMIC-GAN}  & \color{red}69.48   & \color{red}68.25   & \color{red}75.12   & \color{red}39.09   & \color{blue}46.12  & \color{red}23.55  & \color{red}23.76   & \color{red}28.03   & \color{red}11.10   & \color{red}81.97   & \color{red}77.22   & \color{red}72.56
        \\ \bottomrule
\end{tabular}}
\end{table*}

\subsection {Performance with Different Momentum}
\begin{figure}[H]
\centering
\includegraphics[width=0.9\linewidth]{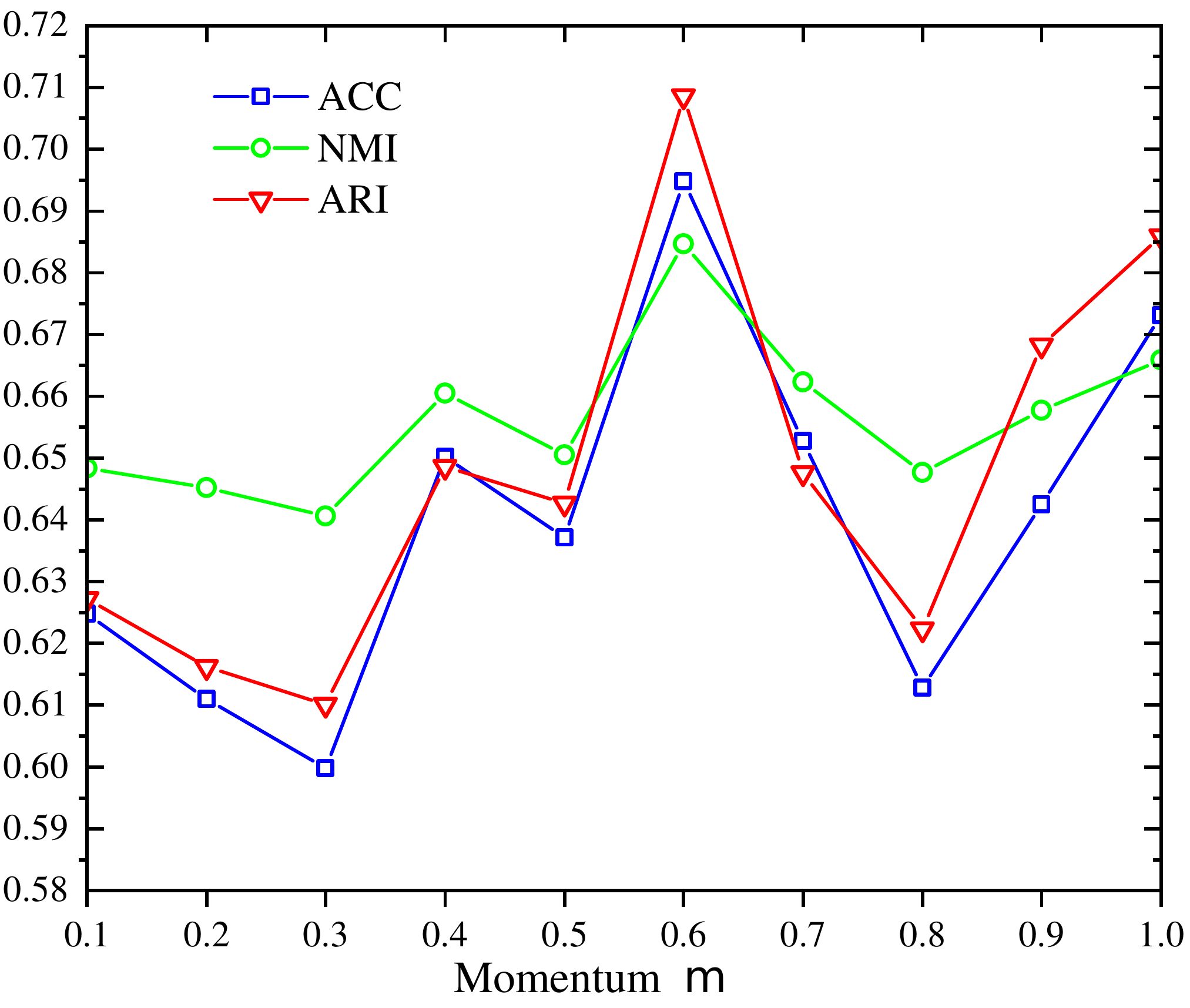}
\caption{\textcolor{black}{Clustering performance on Caltech101-20 with different momentum $m$.}}
\label{fig7}
\end{figure}

In the contrastive prediction module, if the parameters of the online network updated by the current step are directly copied to the target network, the target network's training state will eventually be completely disordered, causing the incapacity to train to oscillate. Therefore, the coordination of states by introducing momentum $m$ is a natural solution. \textcolor{black}{Consistency and complementarity represent the same semantics and different semantics of different views, respectively. If the network is updated faster, the learned representation will contain more consistency, otherwise, it can retain more complementary information. So the introduction of momentum $m$ can also help us find the most suitable balance point.} We designed a set of experiments on the Caltech101-20 dataset with a missing rate (MR) of 0.5. As shown in Fig. \ref{fig7}, the clustering effect is observed by adjusting different values of the momentum $m$.

\textcolor{black}{From the results obtained in Fig. \ref{fig7}, it can be seen that: 1) When the value of $m$ is small, the parameters of the target network change too quickly, ignoring the complementarity between hidden representations. 2) When $m$ is large ($m>0.6$), the gradient update is slow the consistent learning is hindered, and the accuracy will experience a significant drop again. Therefore, too fast or too slow parameter changes are not conducive to consistent learning, because the consistency and complementarity of different views are not balanced. So choosing an appropriate momentum as a balance point can promote the consistent learning of latent representations from different views. In the update of the target network, 60\% of the parameters keep the same value, and the clustering performance is optimal at this time.}

\subsection {Comparisons with state of the arts}\label{sec:comparisons}

For methods that can only handle complete view data, we fill in the missing data with the mean of the same view. For a fair comparison, we use the default network structure and parameters for all methods. We test all methods with an MR of equal to 0.5. Experimental results on the above \textcolor{black}{four} databases are listed in Table \ref{table2}. Compared with the ten clustering methods mentioned above, our CIMIC-GAN achieves state-of-the-art clustering performance. We can observe the following from the results: 1) On all \textcolor{black}{four} datasets, CIMIC-GAN outperforms other methods. 2) In most cases, DAIMC, PVC, EERIMVC, and PIC obtain worse IMC performance than the other methods. These methods all consider consistency learning and data recovery as two separate parts. \textcolor{black}{The contrastive consistency module in CIMIC-GAN strengthens consistency learning, and the contrastive prediction module alleviates inconsistency.} There is a trade-off between these two parts. We can conclude that using double contrastive learning to simultaneously mine complementary and consistent information from multiple perspectives can improve clustering performance. 3) Methods such as DCCAE, PVC, BMVC, and PIC only learn consistency, while DAIMC learns representations that only consider complementary information. Thus, CIMIC-GAN far outperforms the above methods. This shows that effectively utilizing the complementary and consistent information of images can improve clustering performance. 4) CIMIC-GAN outperforms methods with incomplete data that do not participate in training, such as UEAF, DCCA, PIC, COMPLETER, etc. This shows that it is necessary for us to use GAN to mine the hidden information in incomplete data.

\begin{figure*}[t]
\centering
\subfigure[\textcolor{black}{Parameter analysis of ACC}]{
\includegraphics[width=5.5cm]{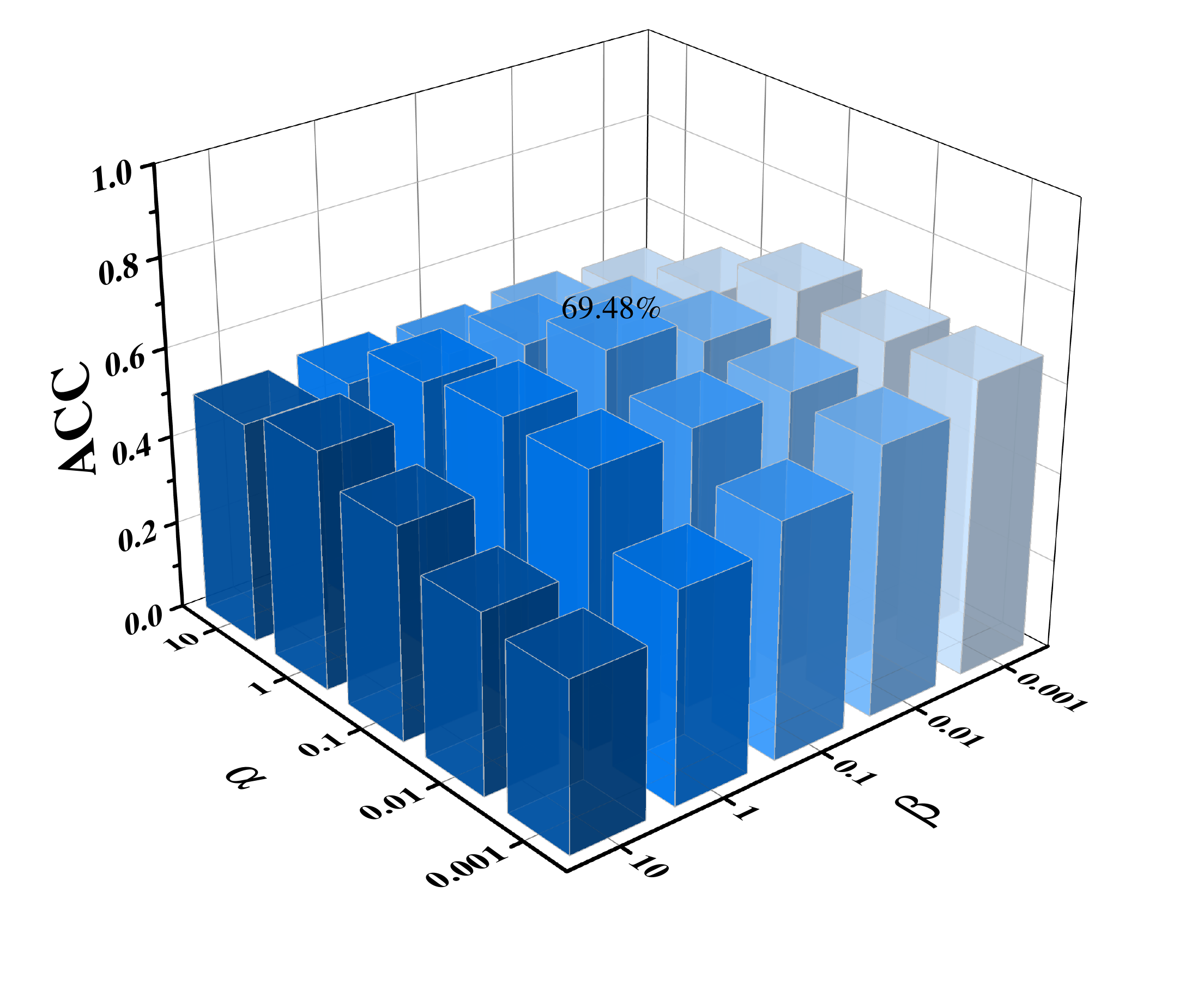}
%\caption{fig1}
}
\quad
\subfigure[\textcolor{black}{Parameter analysis of NMI}]{
\includegraphics[width=5.5cm]{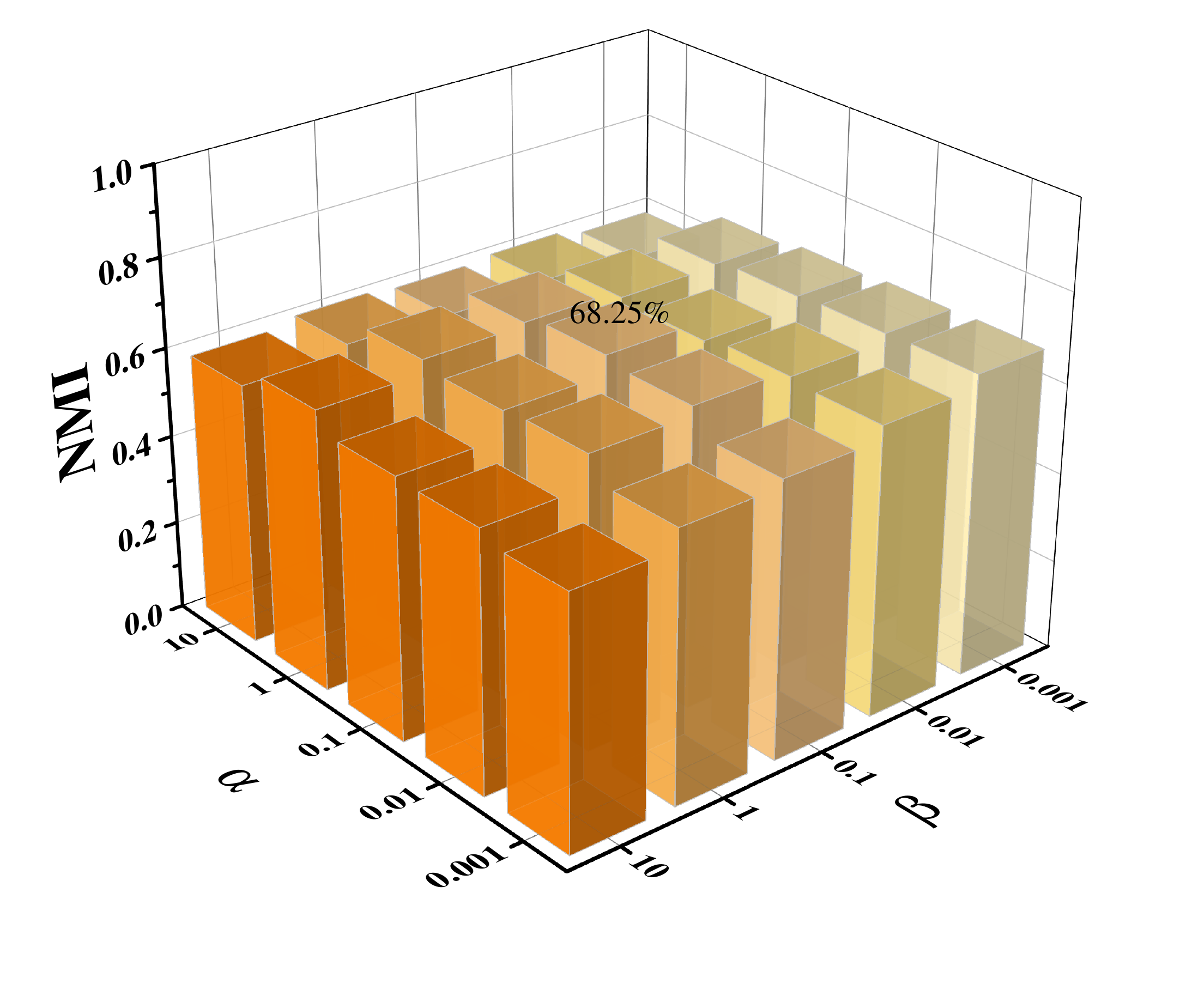}
}
\quad
\subfigure[\textcolor{black}{Parameter analysis of ARI}]{
\includegraphics[width=5.5cm]{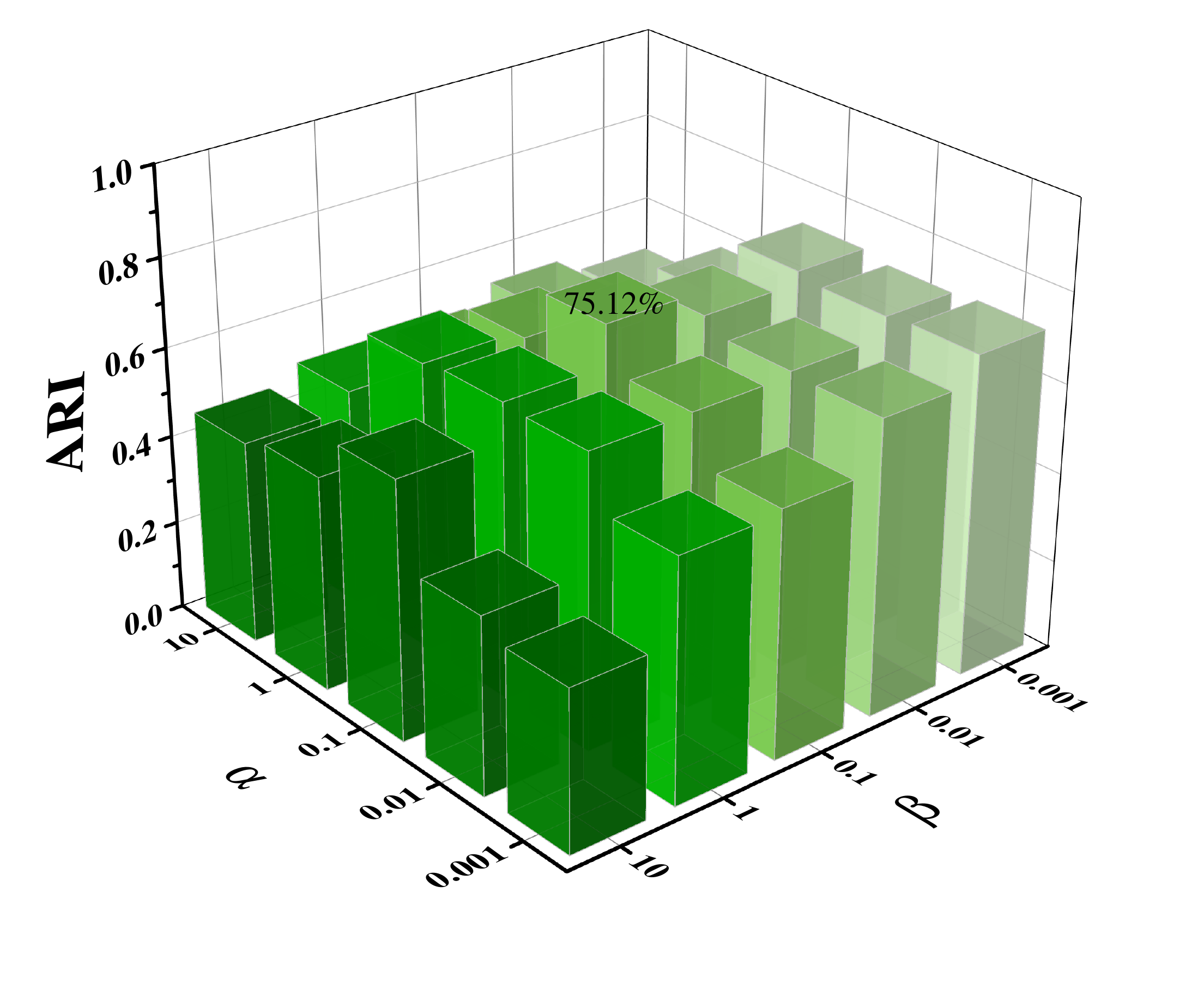}
}
\caption{\textcolor{black}{The clustering performance of CIMIC-GAN on the Caltech101-20 dataset with different parameters $\alpha$ and $\beta$, with a missing rate of 0.5.}}
\label{fig8}
\end{figure*}

\begin{figure*}[htbp]
\centering
\subfigure[Epoch 1 (NMI = 0.4387)]{
\includegraphics[width=4cm]{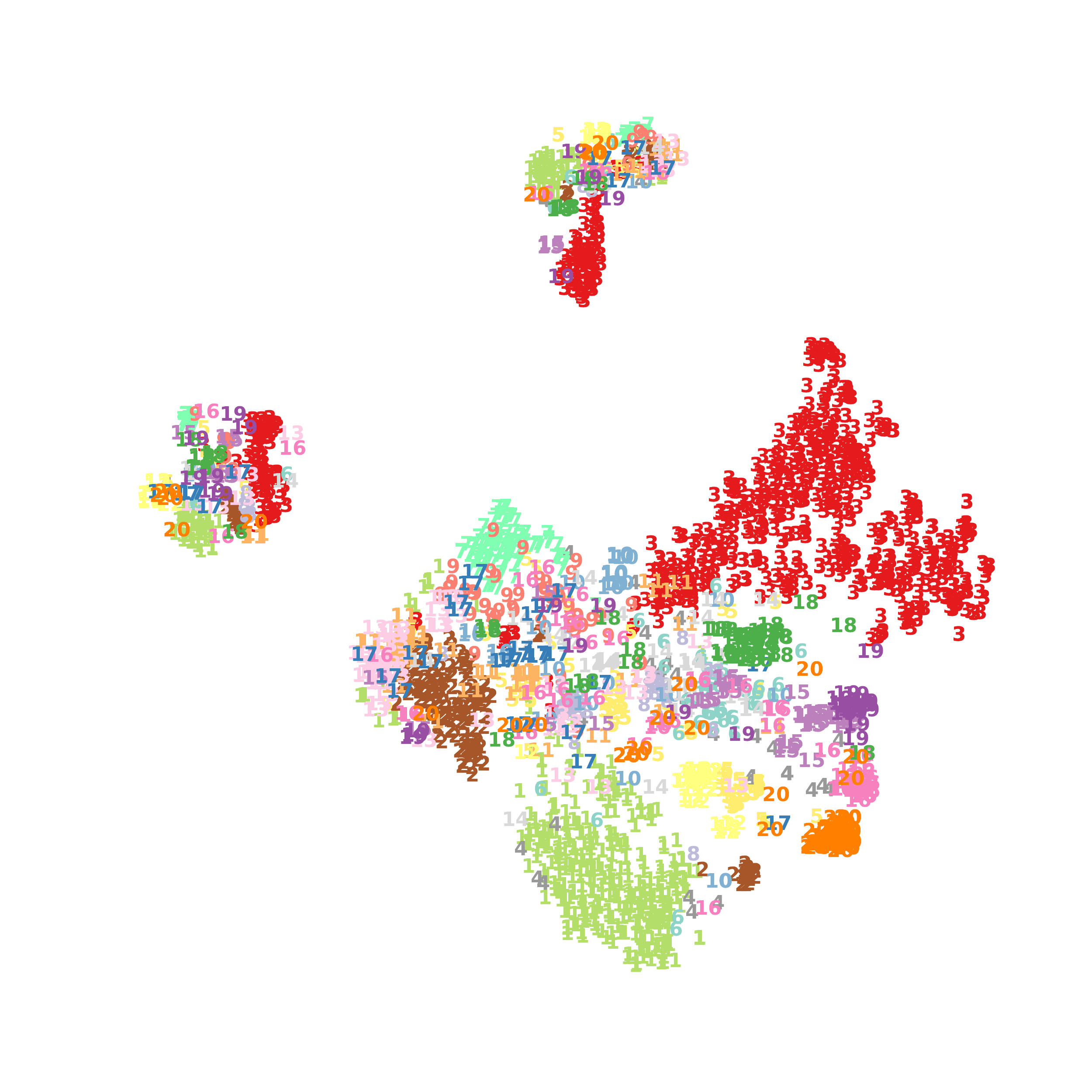}
%\caption{fig1}
}
\quad
\subfigure[Epoch 10 (NMI = 0.4555) ]{
\includegraphics[width=4cm]{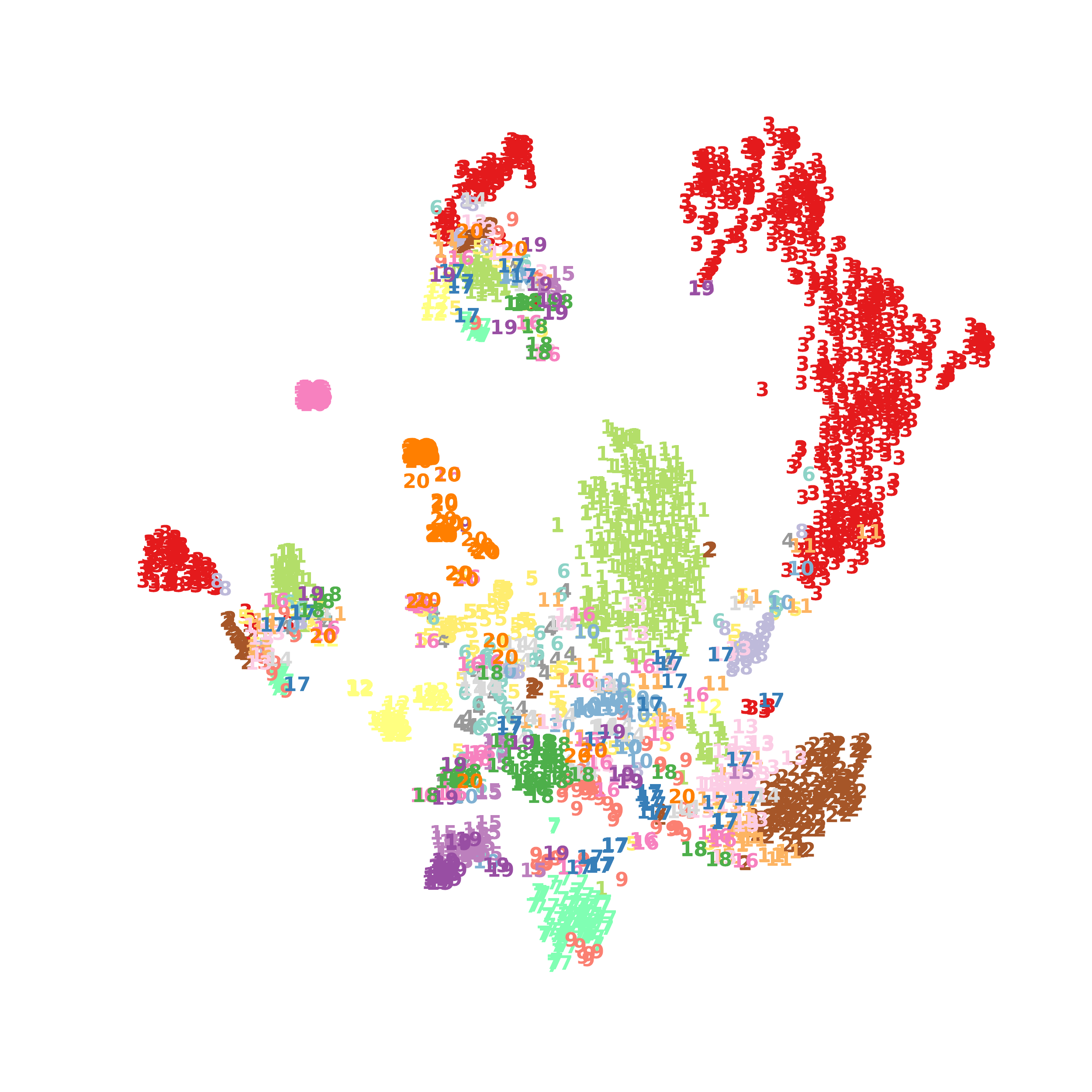}
}
\quad
\subfigure[Epoch 100 (NMI = 0.6767)]{
\includegraphics[width=4cm]{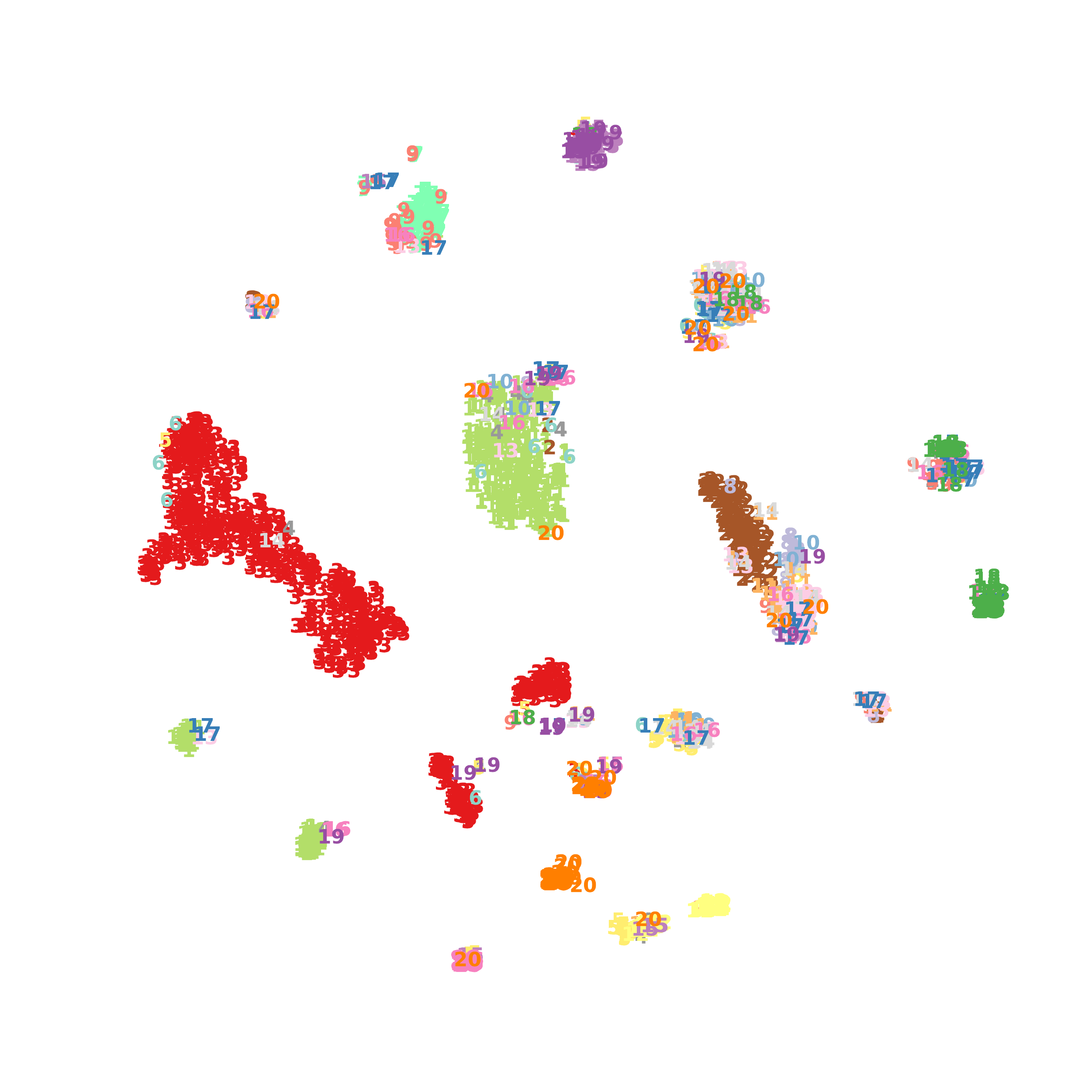}
}
\quad
\subfigure[Epoch 200 (NMI = 0.718)]{
\includegraphics[width=4cm]{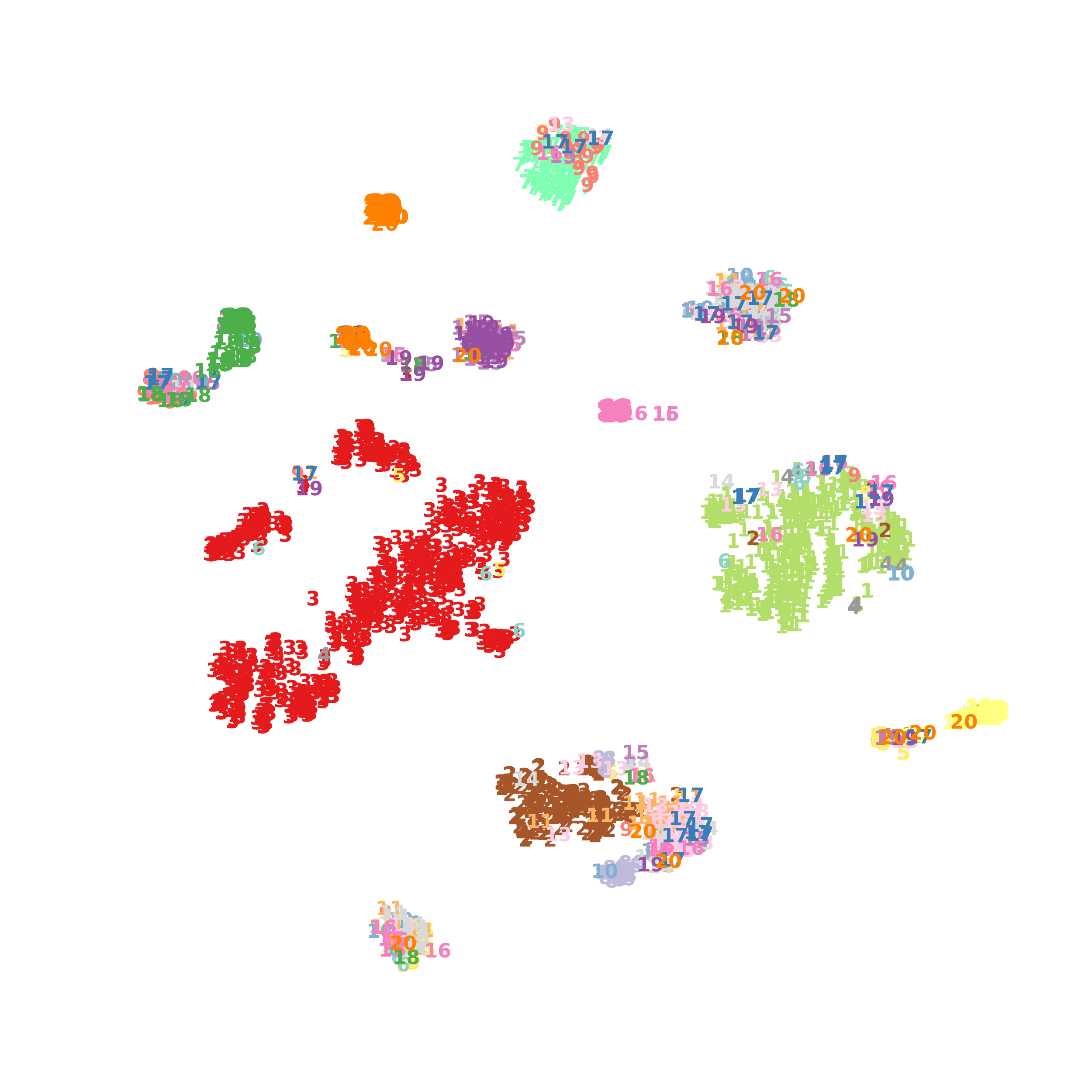}
}
\quad
\subfigure[Epoch 1 (NMI = 0.2314)]{
\includegraphics[width=4cm]{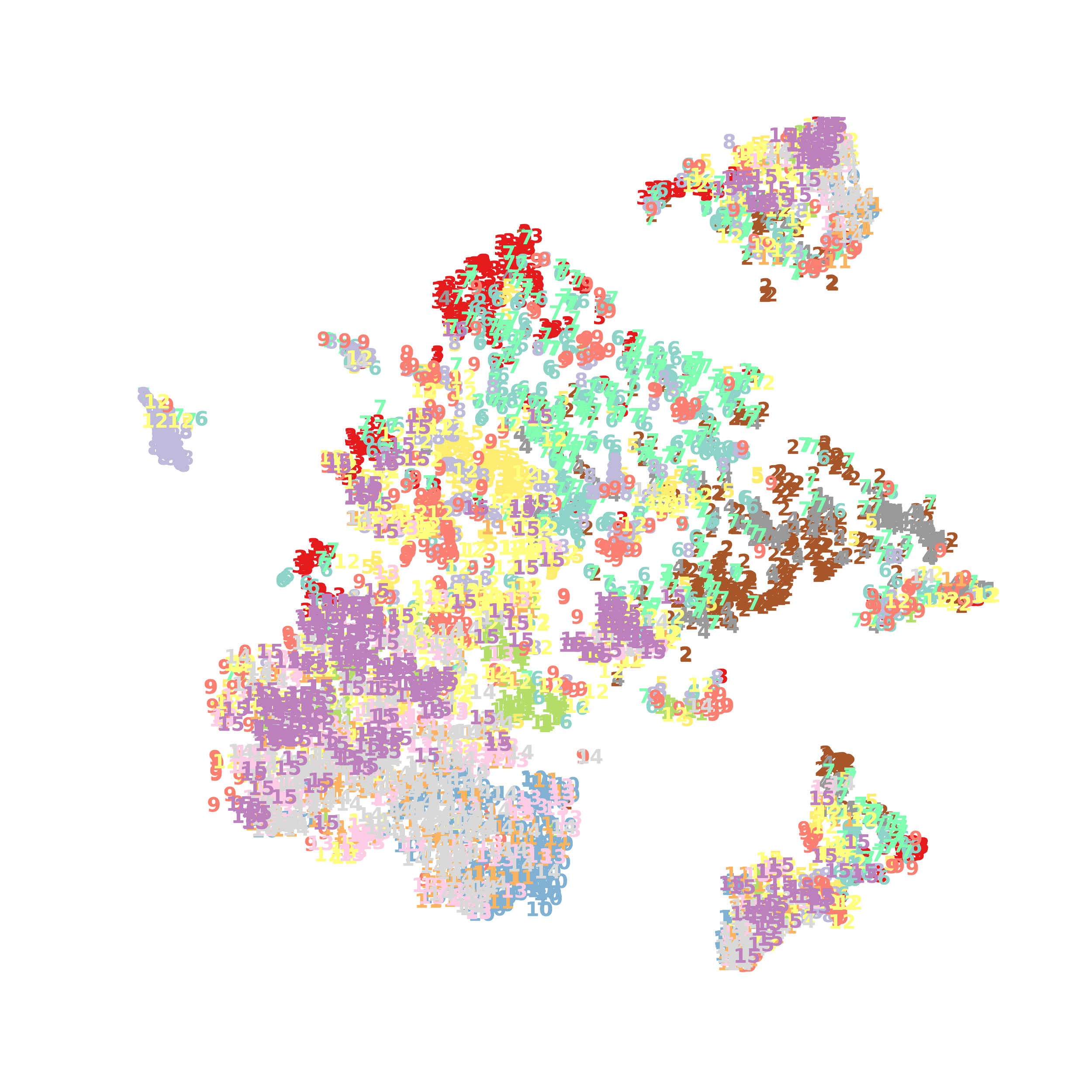}
}
\quad
\subfigure[Epoch 10 (NMI = 0.3465)]{
\includegraphics[width=4cm]{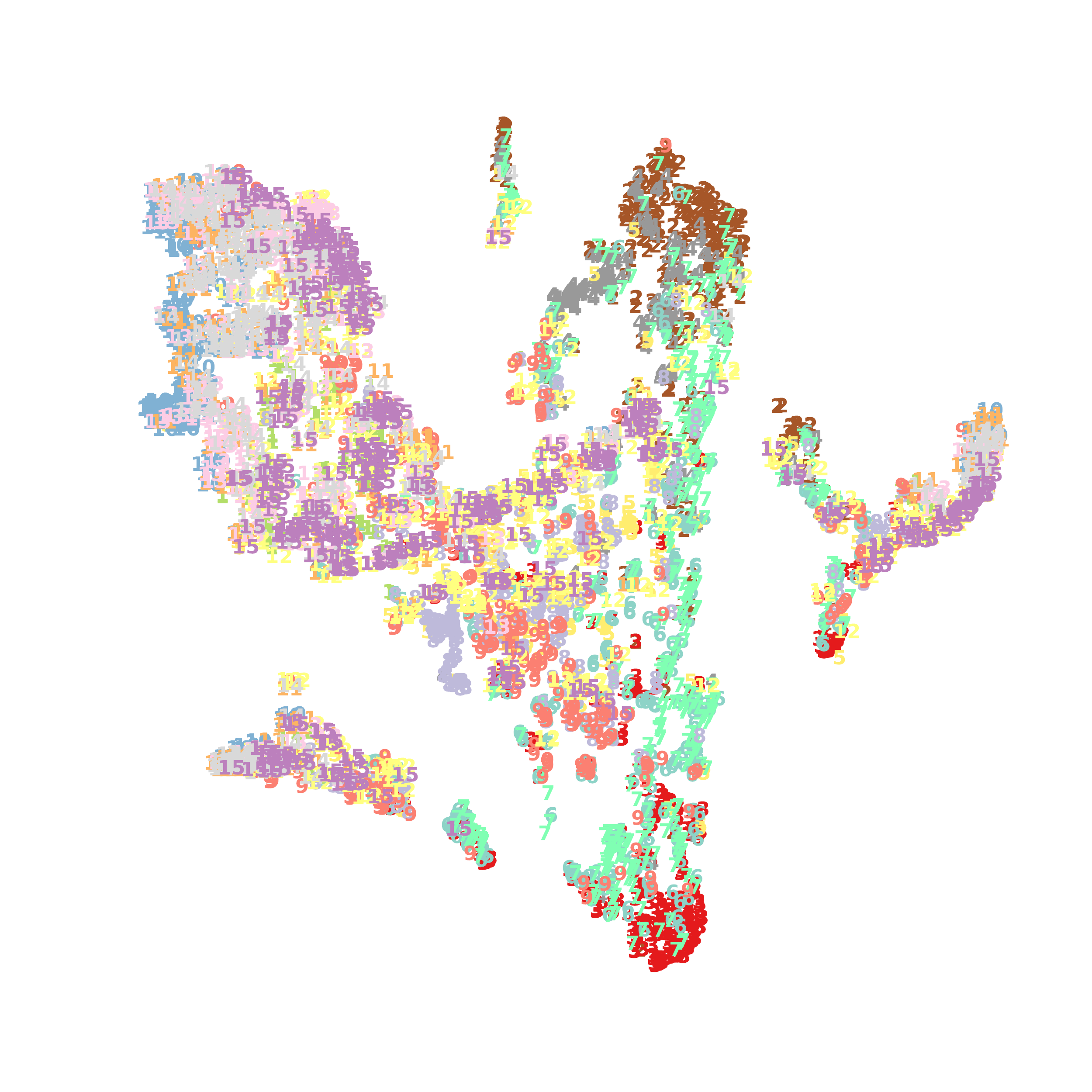}
}
\quad
\subfigure[Epoch 100 (NMI = 0.3997)]{
\includegraphics[width=4cm]{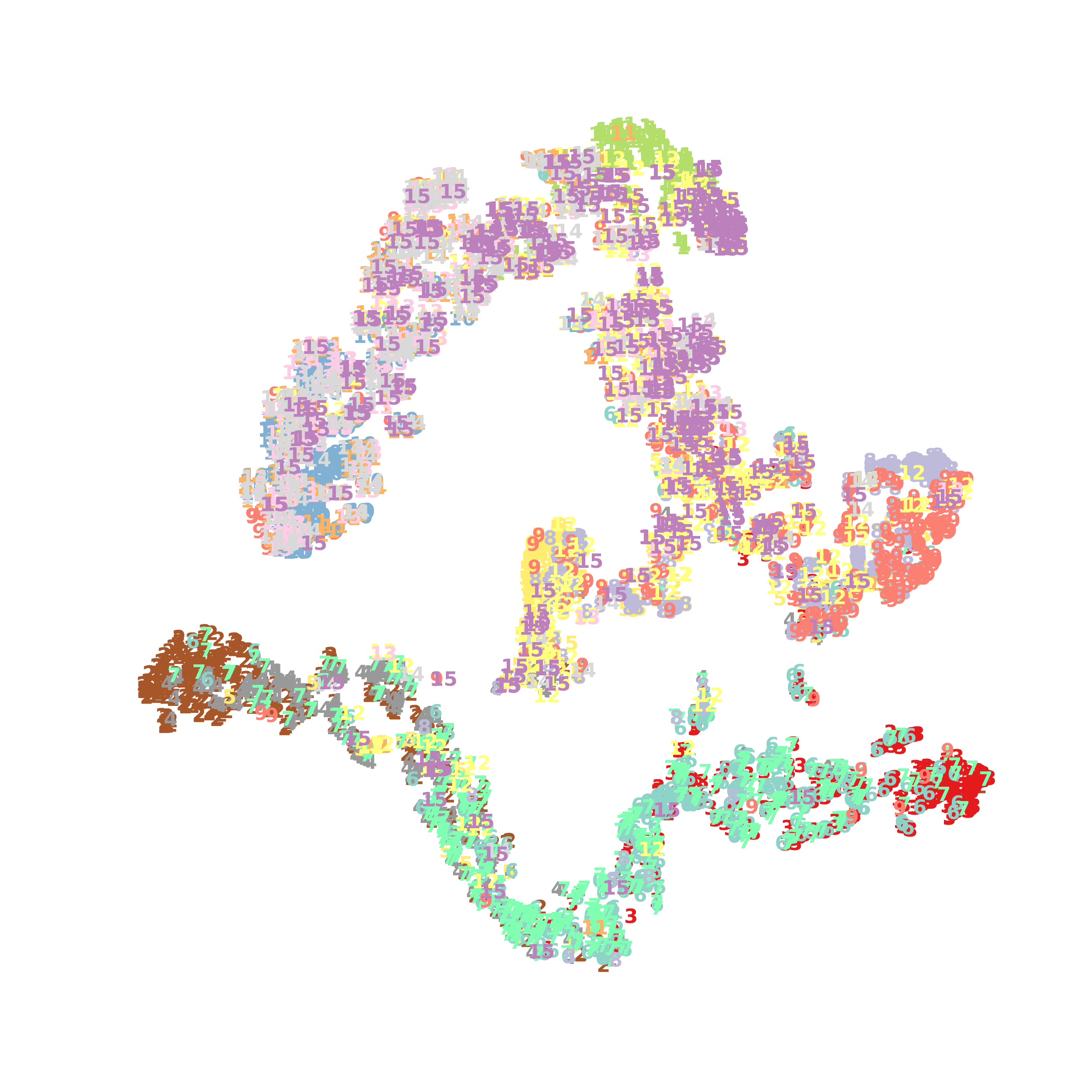}
}
\quad
\subfigure[Epoch 200 (NMI = 0.4600)]{
\includegraphics[width=4cm]{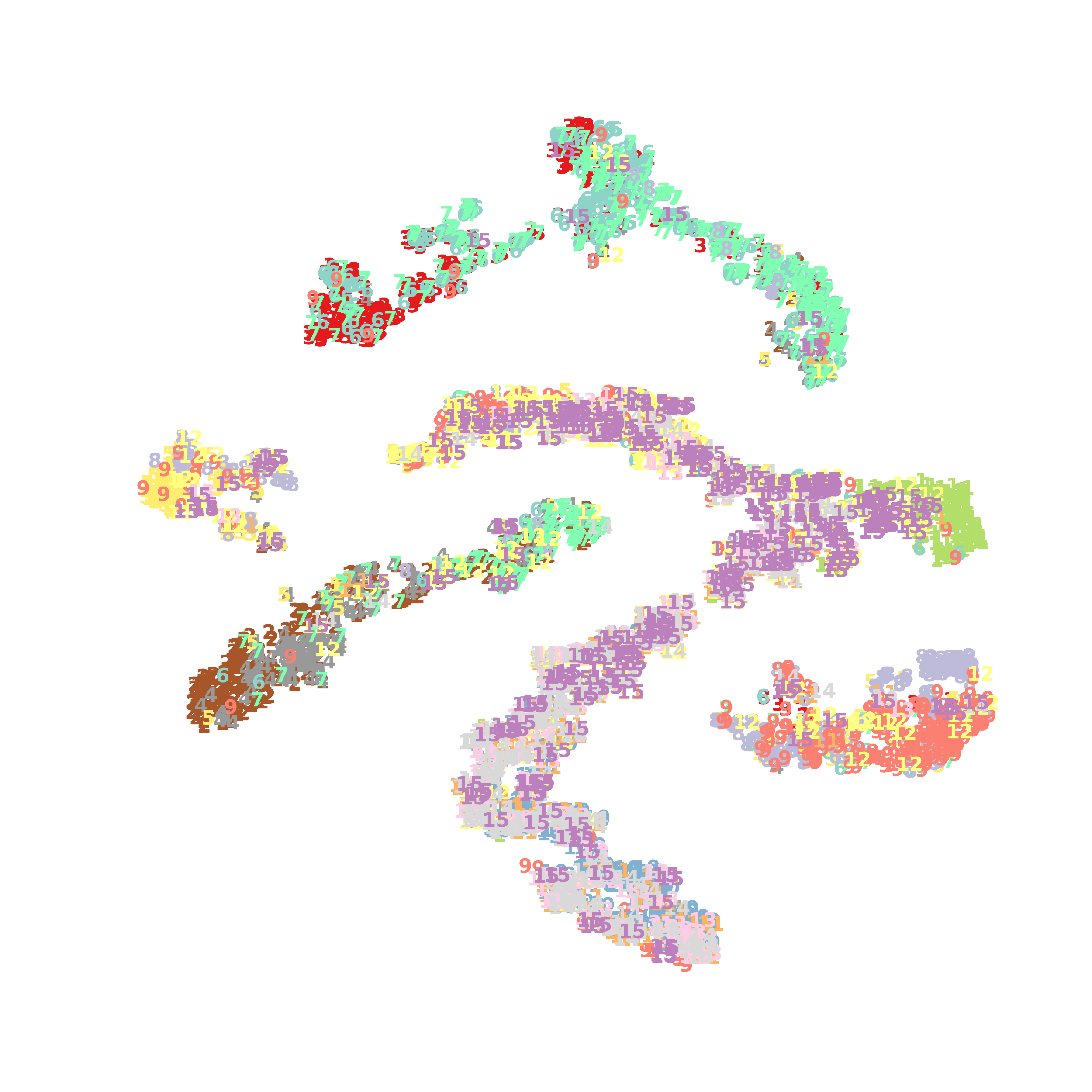}
}
\quad
\subfigure[Epoch 1 (NMI = 0.1818)]{
\includegraphics[width=4cm]{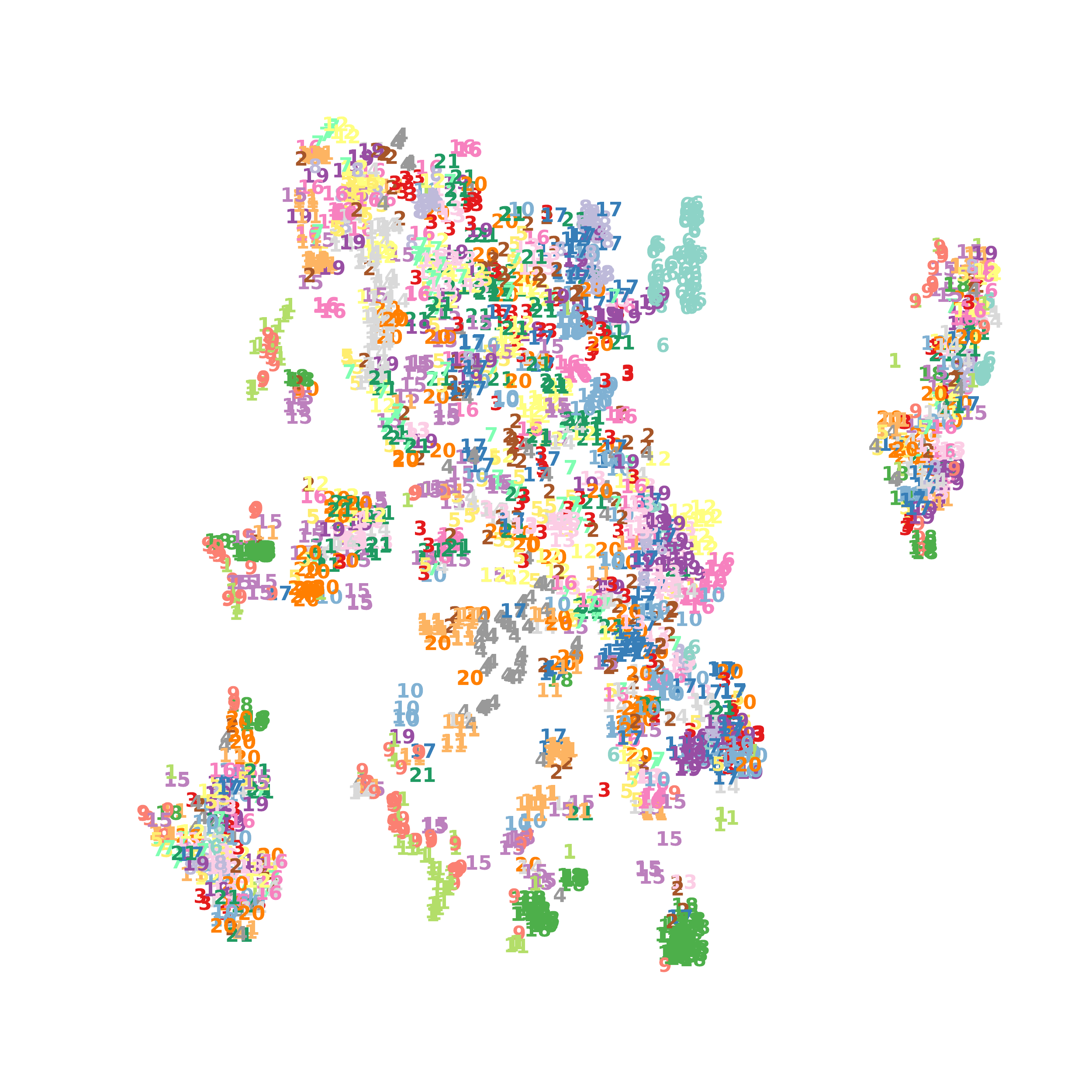}
}
\quad
\subfigure[Epoch 10 (NMI = 0.2228)]{
\includegraphics[width=4cm]{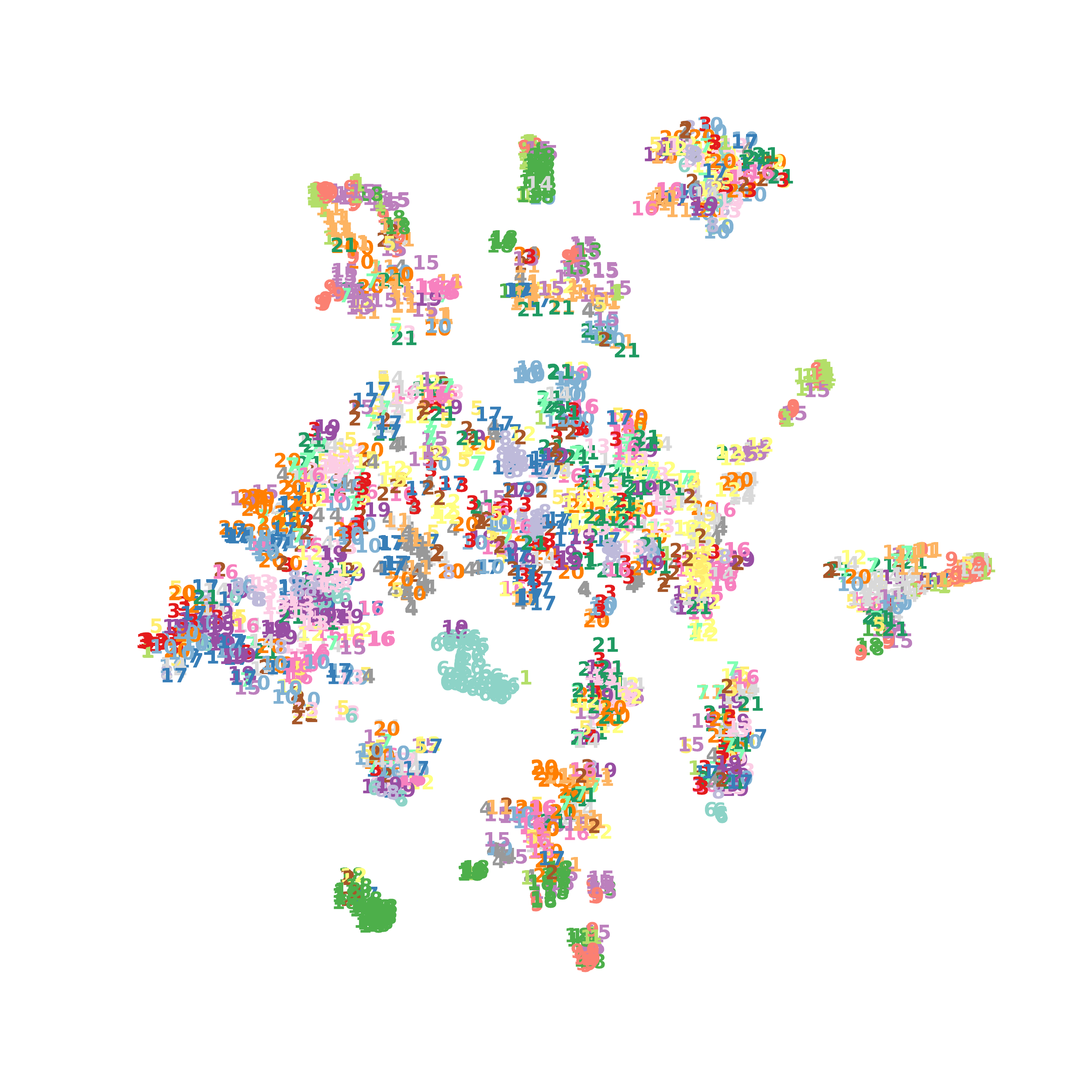}
}
\quad
\subfigure[Epoch 100 (NMI = 0.2887)]{
\includegraphics[width=4cm]{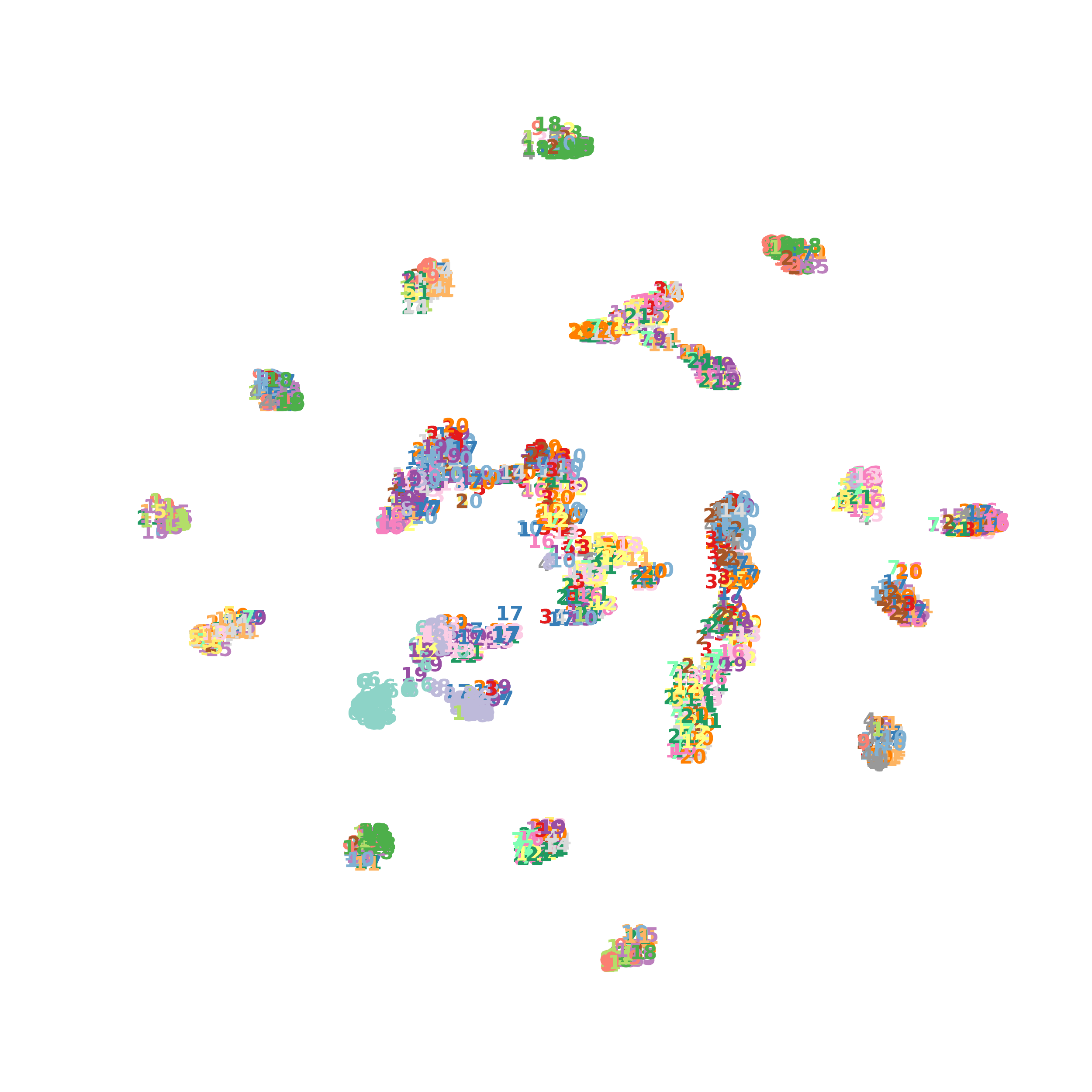}
}
\quad
\subfigure[Epoch 200 (NMI = 0.3004)]{
\includegraphics[width=4cm]{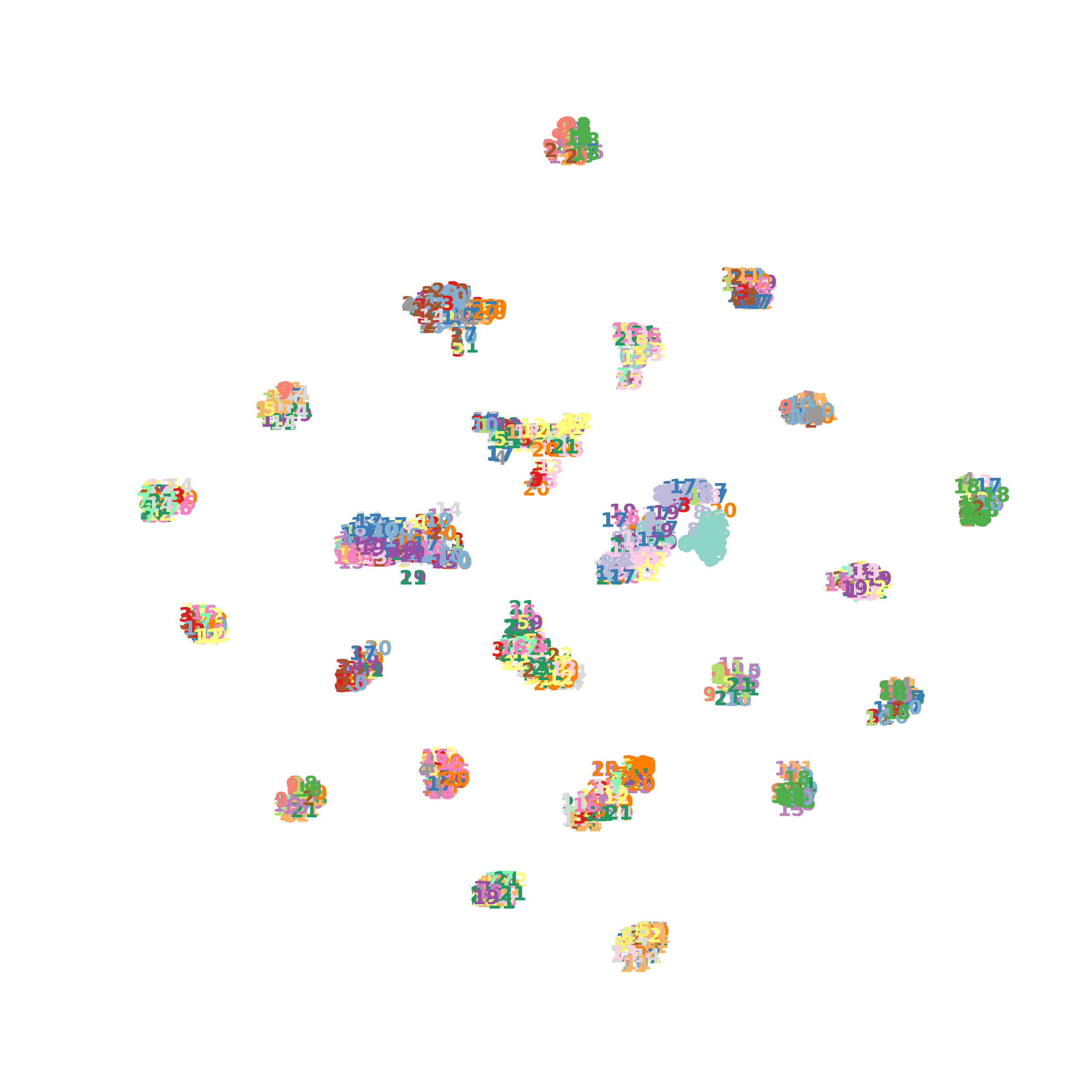}
}
\quad
\subfigure[\textcolor{black}{Epoch 1 (NMI = 0.1278)}]{
\includegraphics[width=4cm]{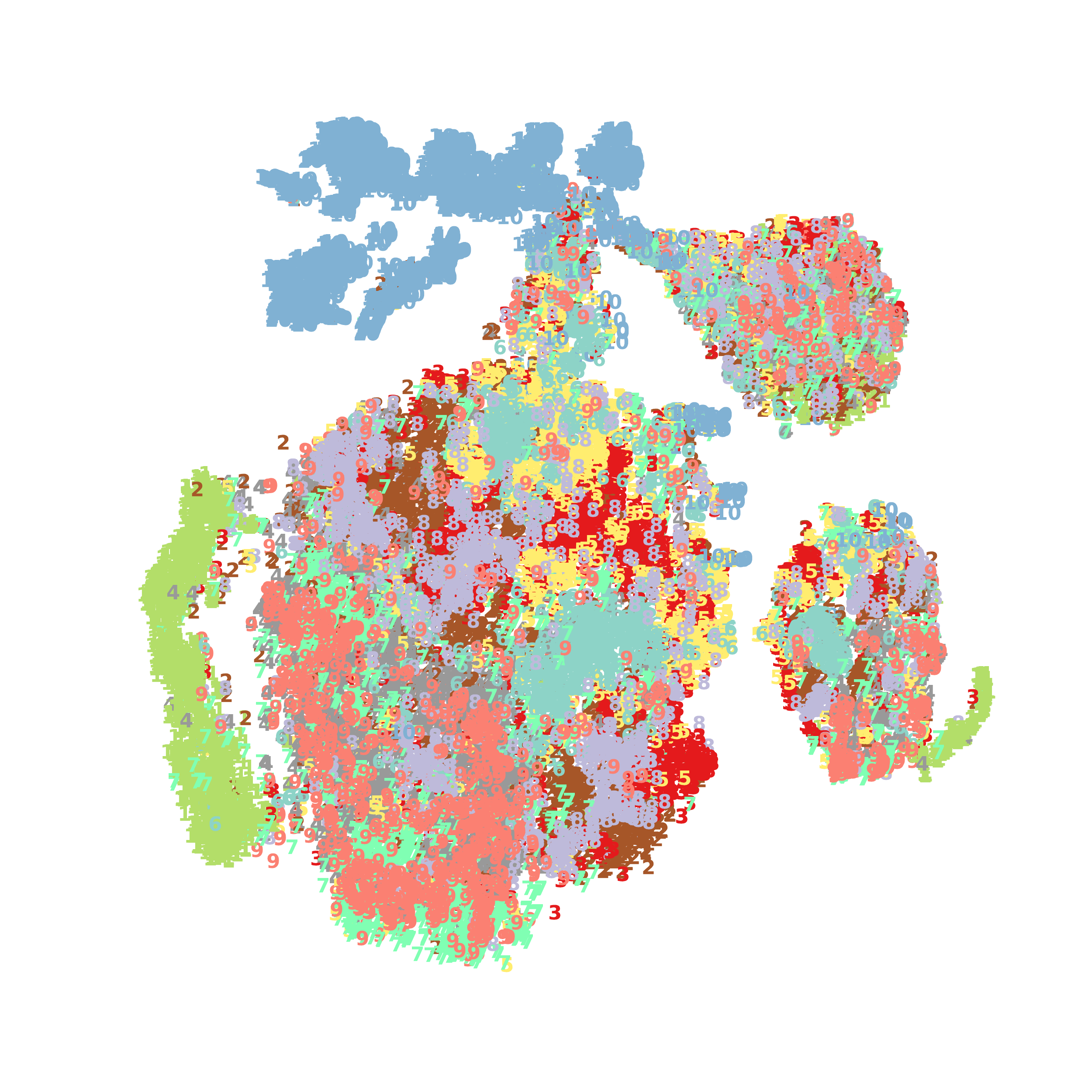}
}
\quad
\subfigure[\textcolor{black}{Epoch 10 (NMI = 0.5555)}]{
\includegraphics[width=4cm]{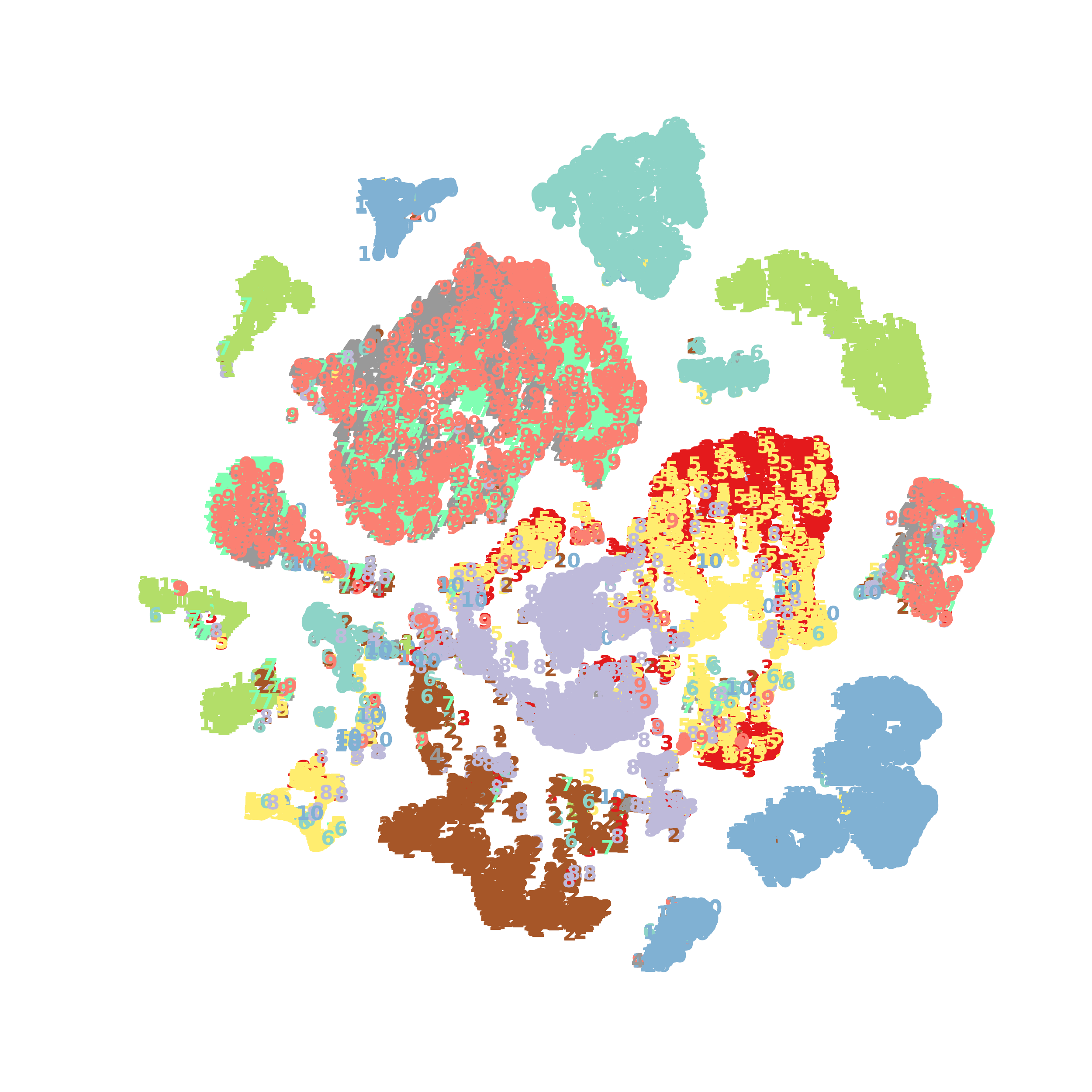}
}
\quad
\subfigure[\textcolor{black}{Epoch 100 (NMI = 0.7222)}]{
\includegraphics[width=4cm]{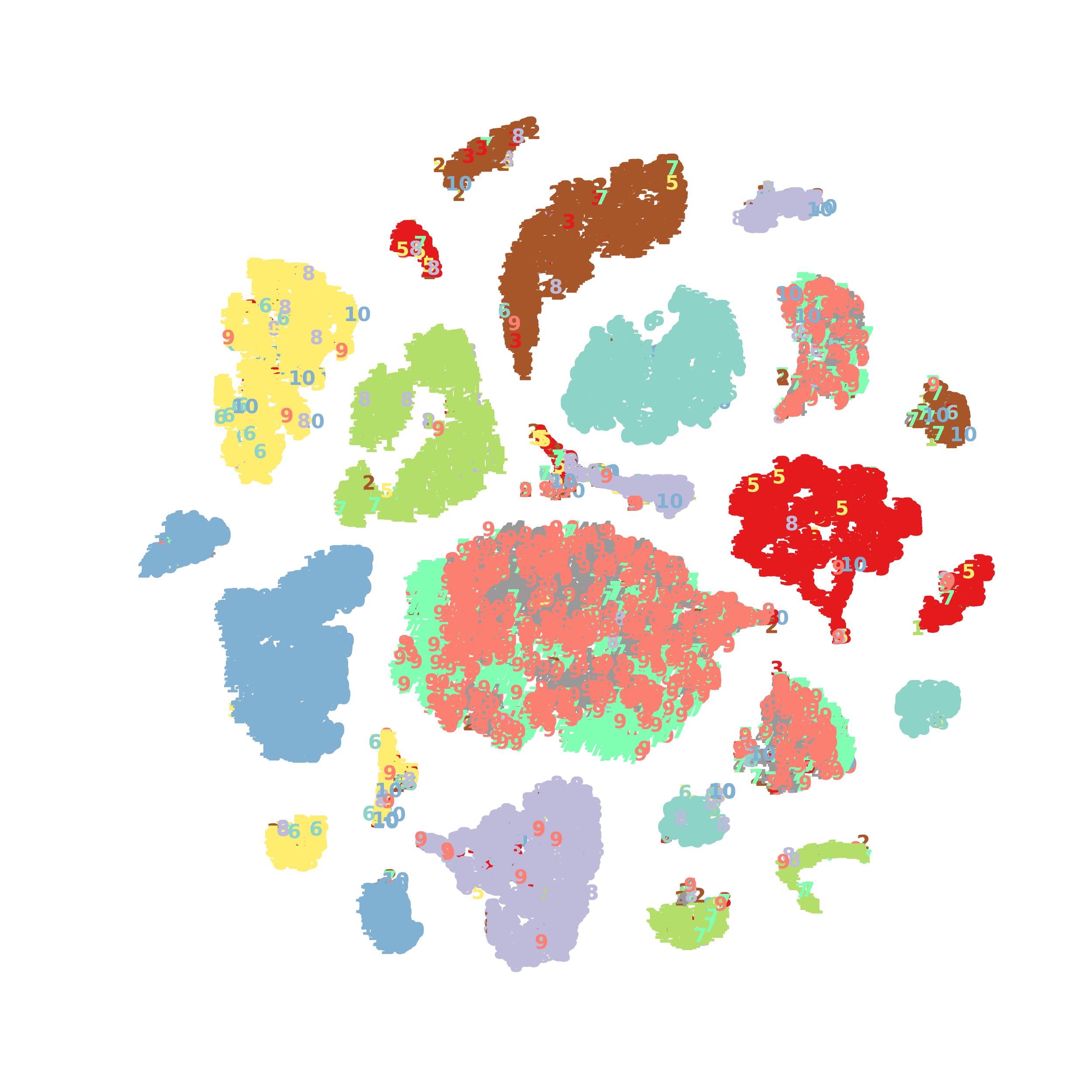}
}
\quad
\subfigure[\textcolor{black}{Epoch 200 (NMI = 0.7722)}]{
\includegraphics[width=4cm]{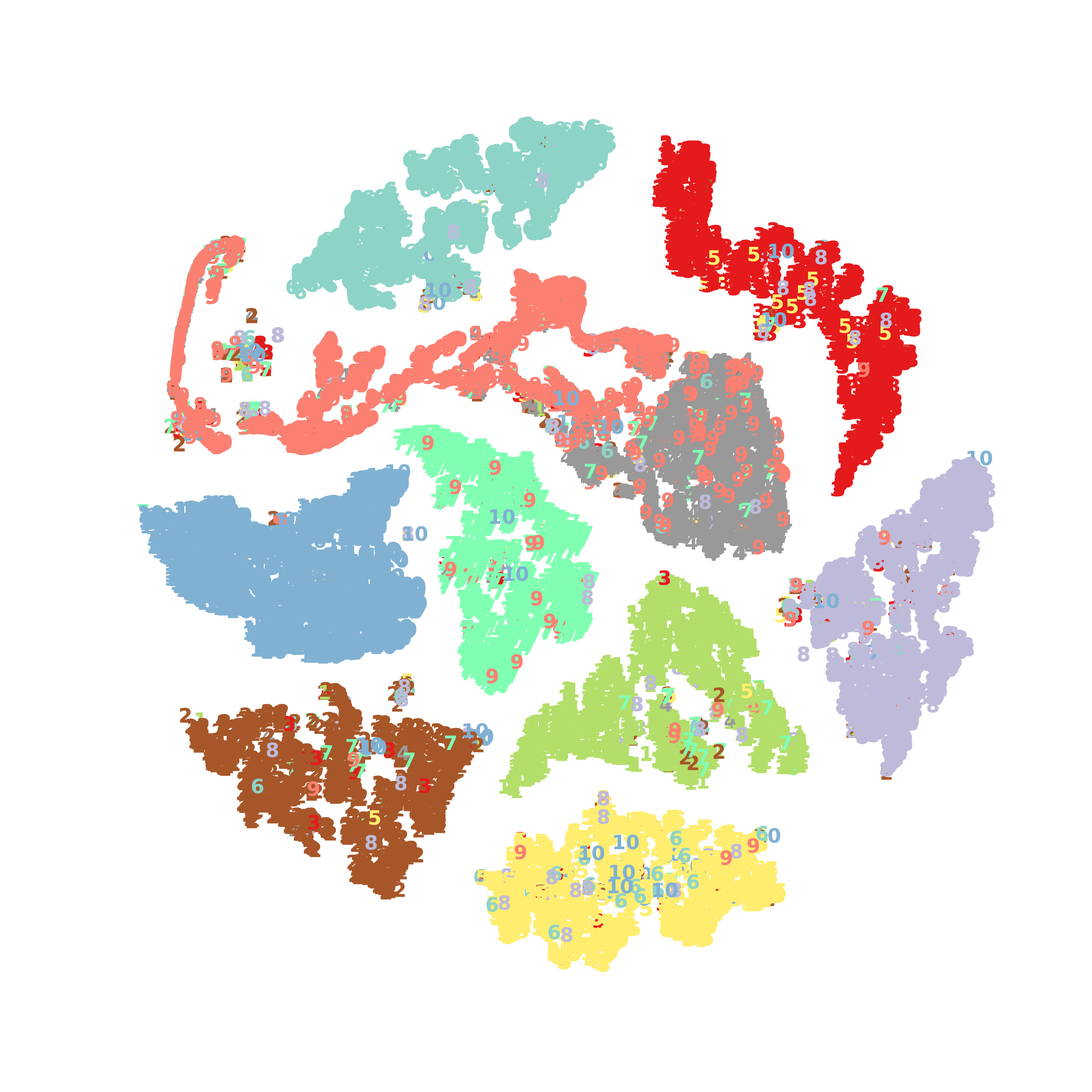}
}

\caption{ Visualization of common repredentations during training. In detail, the figure shows the visualization results of the  representations of Caltech101-20 (a-d), Scene-15 (e-h), LandUse-21 (i-l)  and \textcolor{black}{Noisy MNIST (m-p)} datasets when the epochs are 1, 10, 100, and 200, respectively.}
\label{fig9}
\end{figure*}

\subsection {\textcolor{black}{Performance on different number of views}}

\textcolor{black}{To demonstrate that our method can in principle be extended to more than two views, we implement a set of comparative experiments on different number of views. A dataset Fashion \cite{69} with three views is introduced here to support this set of experiments. %We follow the literature \cite{70} to treat different three styles as three views of one product. 
We leverage this dataset to construct a dataset only with two views, Fashion-2V, whereas we construct Fashion-3V with three views that is missing two views, and Fashion-3V* with three views that is missing one view. See Table \ref{table4} for more detailed settings of these three datasets.}

\begin{table}[htbp]
\renewcommand\arraystretch{1.3}
\centering
\caption{Summary of datasets with different numbers of views.}
\label{table4}
\setlength{\tabcolsep}{1.8mm}{
\begin{tabular}{lcccc}

\toprule \text Datasets     & Size   & Views   & \multicolumn{1}{l}{\# of categories} & Dimension     \\
\hline \text Fashion-2V      & 10,000   &2   & 10                                  & 784/784 \\
             Fashion-3V        & 10,000  &3    & 10                                  & 784/784/784  \\
             Fashion-3V*      &10,000    &3   &10                                   & 784/784/784 \\
             
\bottomrule  \text

\end{tabular}
}
\end{table}

\begin{figure}[H]
\centering
\includegraphics[width=0.95\linewidth]{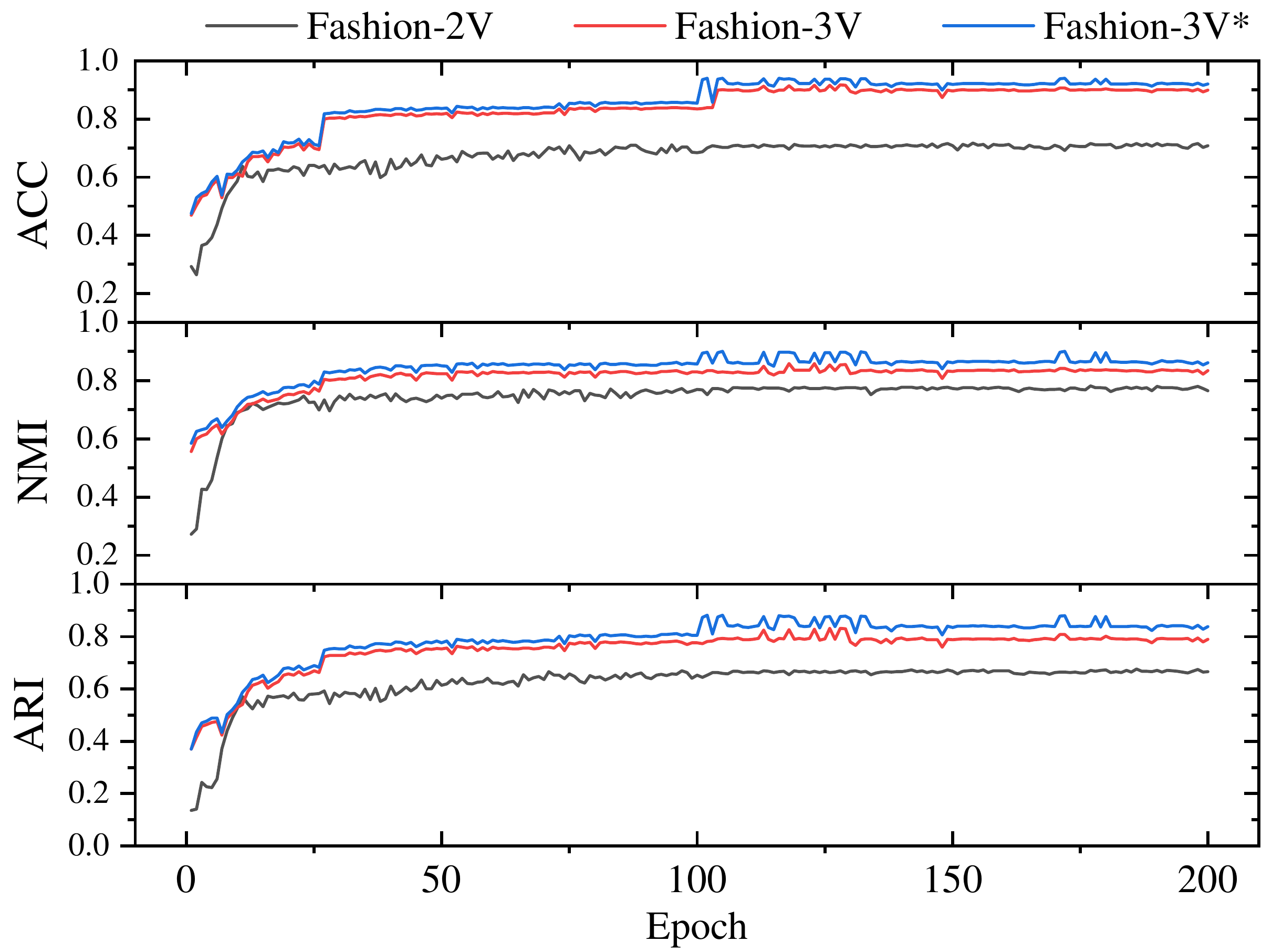}
\caption{\textcolor{black}{Clustering performance on Caltech101-20 with different number of views.} }
\label{F}
\end{figure}

\textcolor{black}{To handle instances with more than two views, CIMIC-GAN needs some extensions. Specifically, we involve another set of network2 to learn the latent representation $Z^{3}$ for the third view. Similarly, we need to introduce two additional sets of prediction networks to realize the mutual prediction of $Z^{1}$ and $Z^{3}$, and $Z^{2}$ and $Z^{3}$. Correspondingly, we maximize the mutual information between the latent representations of any two of these three views (Formula \ref{eq12}). Fig. \ref{F} shows the results of training on Fashion-2V, Fashion-3V and Fashion-3V* for 200 epochs. As the number of instance views changed from two to three, more features are involved and each evaluation metric improved significantly. Among them, the performance of Fashion-3V* is slightly higher than Fashion-3V in an all-around way. This is because our method alleviates the negative impact of missing views, so their two curves are very close, which further proves that CIMIC-GAN has strong robustness and adaptability.}

\subsection {Parameter Analysis and Ablation Studies}

In this section, we analyze CIMIC-GAN on the Caltech101-20 dataset from two perspectives, i.e., parameter analysis and ablation studies.

\textcolor{black}{Our method has two user-specified parameters, i.e., the reconstruction loss trade-off parameter $\alpha$, and the contrastive consistency loss trade-off parameter $\beta$. To evaluate the impact of $\alpha$ and $\beta$, we change their value in the range of {0.001, 0.01, 0.1, 1, 10} when training the model on dataset Caltech101-20. As shown in Fig. \ref{fig8}, when $\alpha$ and $\beta$ are set to 0.1 and 0.1, respectively, the experimental results are the best. Moreover, CIMIC-GAN was run with the same hyper-parameter setting on the \textcolor{black}{four} datasets and achieved good performance (see Section \ref{sec:comparisons}), which means CIMIC-GAN is not very sensitive concerning the hyperparameters.}

We conduct ablation studies with four variants of CIMIC-GAN to show the effectiveness of its different modules. (1) We only use the reconstruction module multi-modal clustering. (2) We only use the contrastive prediction module multi-modal clustering. (3) We only use the contrastive consistency module multi-modal clustering. (4-6) We combine the above three modules in pairs. (7) We use the complete Network1 with three modules to learn consistent cluster predictions of multiple modalities. (8) Compared with setting “(7)”, we additionally added GAN in Network2.

% \begin{table}[htbp]
% \renewcommand\arraystretch{1.3}
% \centering
% \caption{ Ablation study on Caltech101-20 with a missing rate of 0.5}
% \label{table3}
% \setlength{\tabcolsep}{2.5mm}{
% \begin{tabular}{lccc}
% \toprule \text moudules of CIMIC-GAN                           & ACC     & NMI    & ARI     \\
% \hline \text $\text{ (1) }\mathcal{L}_{rec}$         & 32.06   & 30.58  & 14.16  \\
%              $\text{ (2) } \mathcal{L}_{pre}$         & 34.94   & 33.22  & 25.02  \\
%              $\text{ (3) } \mathcal{L}_{cc}$          & 46.59   & 57.36  & 41.56  \\
%              $\text{ (4) } \mathcal{L}_{r e c}+\mathcal{L}_{\text {pre}}$        & 53.36   & 43.35  & 27.96  \\
%              $\text{ (5) } \mathcal{L}_{r e c}+\mathcal{L}_{\text {cc}}$         & 55.45   & 62.67  & 54.59  \\
%              $\text{ (6) } \mathcal{L}_{cc}+\mathcal{L}_{\text {pre}}$           & 64.58   & 62.10  & 70.87  \\
%              $\text{ (7) } \mathcal{L}_{r e c}+\mathcal{L}_{c c}+\mathcal{L}_{\text {pre }}$         & 66.32   & 66.30  & 69.69  \\
%              $\text{ (8) } \mathcal{L}_{r e c}+\mathcal{L}_{c c}+\mathcal{L}_{\text {pre }}$ and $\mathcal{L}_{adv}$        & 69.48   & 68.25  & 75.12  \\

% \bottomrule \text

% \end{tabular}
% }
% \end{table}

\begin{table}[htbp]
\renewcommand\arraystretch{1.3}
\centering
\caption{ Ablation study on Caltech101-20 with a missing rate of 0.5}
\label{table3}
\setlength{\tabcolsep}{2.5mm}{
\begin{tabular}{lccc}
\toprule \text moudules of CIMIC-GAN                           & ACC     & NMI    & ARI     \\
\hline \text (1)$\mathcal{L}_{rec}$         & 32.06   & 30.58  & 14.16  \\
             (2)$\mathcal{L}_{pre}$         & 34.94   & 33.22  & 25.02  \\
             (3)$\mathcal{L}_{cc}$          & 46.59   & 57.36  & 41.56  \\
             (4)$\mathcal{L}_{rec}+\mathcal{L}_ {pre}$        & 53.36   & 43.35  & 27.96  \\
             (5)$\mathcal{L}_{rec}+\mathcal{L}_ {cc}$         & 55.45   & 62.67  & 54.59  \\
             (6)$\mathcal{L}_{cc}+\mathcal{L}_{pre}$           & 64.58   & 62.10  & 70.87  \\
             (7)$\mathcal{L}_{rec}+\mathcal{L}_{c c}+\mathcal{L}_{pre}$         & 66.32   & 66.30  & 69.69  \\
             (8)$\mathcal{L}_{rec}+\mathcal{L}_{cc}+\mathcal{L}_{pre} + \mathcal{L}_{adv}$        & 69.48   & 68.25  & 75.12  \\

\bottomrule \text

\end{tabular}
}
\end{table}

Table \ref{table3} shows the loss components and experimental results corresponding to the four variants. In row (2), it can be seen that optimization alone with contrastive prediction loss $\mathcal{L}_{pre}$ may lead to trivial solutions or model collapse because $\mathcal{L}_{rec}$ is not optimized and thus the low-dimensional latent representation loses more complementary information. This also demonstrates that the key task of IMC is to mine complementary information to achieve consistent predictions across multiple modalities. Comparing row (6) with row (4) and (5), dual contrastive learning is more effective than a single consistency learning module. Comparing row (7) with row (1), it can be inferred that double contrast learning has a great improvement in the clustering performance, and shows that the improvement in consistency by the double contrastive learning module is huge. Comparing row (7) with row (8), the introduction of GAN in the encoding process makes the hidden information of incomplete data more fully utilized. Significantly, one can find that each module of CIMIC-GAN has improved the clustering performance, which further proves the effectiveness of our motivations and the proposed framework.

\subsection {Visualization}

As shown in Fig. \ref{fig9}, we show t-sne \cite{55} visualizations of the obtained representations on \textcolor{black}{four} datasets.  In the experiments, the missing rate is fixed to 0.5. The common representation itself is a fusion representation, which fuses a complete view representation and a missing view representation, where the missing view representation is filled by incomplete data through a contrastive prediction module. Therefore, Fig. \ref{fig9} illustrates the learning process of Network1. With the increase of epochs, more advanced semantic information is achieved, the common representations learned by Network1 become more compact and independent, and the density of clusters is higher. The advanced semantic information enables better consistent learning. For all these \textcolor{black}{four} datasets, Network1 converges at around 200 epochs, indicating that our proposed model is very robust.

%% file: Conclusion.tex
\section{Conclusion}
This paper proposes CIMIC-GAN to effectively exploit the hidden information in incomplete data to learn multi-view consistent semantics. \textcolor{black}{CIMIC-GAN works well without paired-view data and achieves optimal clustering performance even in the case of high missing rates. But there are also some disadvantages, such as the model not being easy to expand when the number of views is large. This work aims to provide a general framework for IMC of visual instances, where the models may also suffer from inherent bias in the data, especially in the case of dirty samples. The framework learns feature extractors and predictors, which can be used in areas such as feature compression, unsupervised labeling, and cross-modal feature retrieval. In the future, we plan to further explore the potential of our framework in other multi-view learning tasks, e.g., 3D reconstruction.}